\documentclass[preprint,12pt]{elsarticle}

\usepackage{amssymb}
\usepackage{amsmath}

\usepackage{booktabs}
\usepackage{multirow}
\usepackage{xcolor}
\usepackage{hyperref}
\usepackage{algorithm}
\usepackage{algorithmic}
\usepackage{subcaption}
\usepackage{graphicx}
\usepackage{float}
\usepackage[section]{placeins}
\usepackage{enumitem}
\setlist{itemsep=2pt, topsep=4pt}
\usepackage{rotating}
\usepackage{threeparttable}

  {\par\begingroup\color{red}\noindent\textbf{[REVISE:]}\par}%
  {\par\endgroup}

\journal{Medical Image Analysis}

\begin{document}

\begin{frontmatter}

\title{Multi-Granularity 3D Kidney Lesion Characterization from CT Volumes}

\author[ufhobi]{Renjie Liang, MSc}
\author[ufhobi]{Zhengkang Fan, MSc}
\author[ufhobi]{Jinqian Pan, MSc}
\author[ufhobi]{Chenkun Sun, MSc}
\author[iu,ri]{Jiang Bian, PhD}
\author[ufurology]{Russell Terry, MD}
\author[ufhobi]{Jie Xu, PhD\corref{cor1}}
\cortext[cor1]{Corresponding author. Email: xujie@ufl.edu}

\affiliation[ufhobi]{
    organization={Department of Health Outcomes and Biomedical Informatics, University of Florida},
    addressline={2004 Mowry Road},
    city={Gainesville},
    postcode={32611},
    state={FL},
    country={USA}
}

\affiliation[ufurology]{
    organization={Department of Urology, University of Florida},
    addressline={1600 SW Archer Road},
    city={Gainesville},
    postcode={32611},
    state={FL},
    country={USA}
}

\affiliation[iu]{
    organization={Department of Biostatistics and Health Data Science, Indiana University School of Medicine},
    addressline={340 West 10th Street},
    city={Indianapolis},
    postcode={46202},
    state={IN},
    country={USA}
}

\affiliation[ri]{
    organization={Center of Biomedical Informatics},
    addressline={1101 W 10th St, Indianapolis},
    city={Indianapolis},
    postcode={46202},
    state={IN},
    country={USA}
}

\begin{abstract}
Radiology reports describe kidney lesions by type, size, enhancement, and attenuation, yet existing 3D methods predict only at the patient or organ level. We reformulate kidney CT characterization as a per-lesion set-prediction task: one model emits a variable number of lesions per kidney, each with four clinical attributes. We curated 2,619 CT volumes from 788 patients at one academic medical center, with multi-granularity side- and per-lesion labels, and used KiTS23 (489 cases) for zero-shot external validation. We propose \textbf{LesionDETR}, a DETR-style architecture with size-distance Hungarian matching and a hierarchical loss that aggregates per-slot outputs to side-level objectives. Across four input representations and six encoder initializations, two design choices dominate: a segmentation mask as an input channel, and same-domain abdominal pretraining (SuPreM); generic large-corpus pretraining is no better than random initialization. LesionDETR reaches bilateral side-level abnormality AUC $0.799 \pm 0.009$ on UF-Health and $0.817 \pm 0.072$ on KiTS23. A count-conditioned variant reaches per-lesion mAP $0.190 \pm 0.083$ on cystic lesions; rare solid-lesion AP stays at the noise floor, pointing to targeted data collection, not architecture, as the next bottleneck. The framework yields verified per-lesion predictions for downstream structured report generation.
\end{abstract}


\end{frontmatter}

\section{Introduction}
\label{sec:introduction}

Computed tomography (CT) is the primary imaging modality for evaluating kidney lesions, which range from benign cysts to renal cell carcinoma~\cite{silverman2019bosniak,herts2018acr}. With over 434,000 new renal cancer cases globally each year~\cite{sung2021global} and increasing utilization of CT in clinical practice, the demand for accurate and consistent interpretation continues to grow. In routine radiology workflow, kidney lesions are described in a lesion-centric manner: each lesion is individually characterized by its location (left or right kidney), type (e.g., cyst or solid mass), size, and imaging features such as enhancement and attenuation. Automating this lesion-centric characterization could benefit multiple stakeholders: radiologists through pre-populated structured reports and reduced risk of missed findings~\cite{liang2025kidney}, patients through more consistent and timely diagnoses, and healthcare systems through improved throughput and standardized documentation.

Existing computational methods, however, do not match this lesion-centric reporting approach. Most prior work operates at the patient or organ level, producing a single prediction per case, such as tumor subtype classification or benign versus malignant diagnosis~\cite{zhou2021endtoend,lacpanet2024,han2019classification}. While segmentation methods for kidney structures from 3D CT have achieved high accuracy~\cite{isensee2021nnunet,wasserthal2023totalseg,myronenko2023kits}, they primarily provide spatial boundaries and do not capture the clinical attributes required in radiology reports, particularly for lesion types where the agreement between segmentation outputs and clinical reports is limited. Even approaches that localize lesions do not model the structured, multi-attribute descriptions required for clinical interpretation. 
Consequently, existing paradigms fail to capture the nuanced semantics—such as enhancement and attenuation—that are diagnostic cornerstones.
Lesion-centric characterization requires both a suitable model and appropriate data. Unlike conventional classification or segmentation tasks, the model must perform set-prediction to handle a variable number of lesions, each associated with multiple clinical attributes. It must also extract imaging features directly from 3D CT volumes, as attributes such as enhancement patterns cannot be fully captured by spatial masks alone. On the data side, lesion-level attribute annotations are not available in existing public datasets and must be derived from clinical reports.

\begin{figure}[!htbp]
    \centering
    \includegraphics[width=\textwidth]{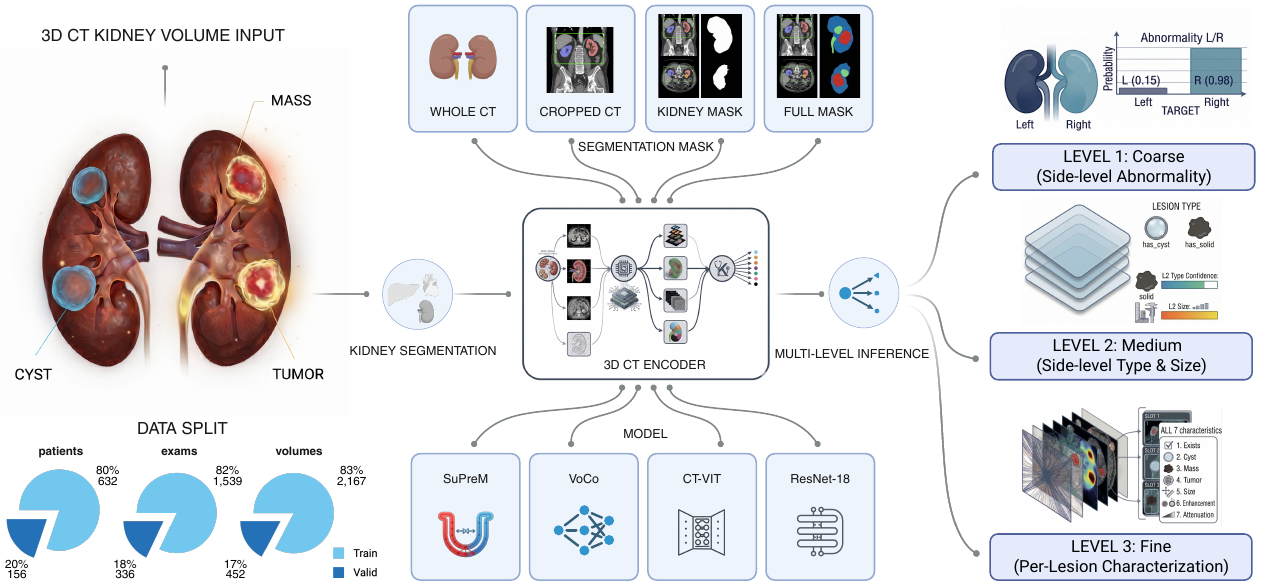}
    \caption{Overview of the lesion-centric characterization framework. A 3D kidney CT volume and its segmentation mask are combined into one of several input representations, encoded by a 3D CT encoder, and passed to multi-level prediction heads outputting side-level abnormality (L1), side-level lesion type and size (L2), and per-lesion attributes (L3). A hierarchical supervision loss jointly trains all three levels from the per-lesion outputs.}
    \label{fig:pipeline}
\end{figure}

We formulate lesion-centric characterization as a prediction task over 3D kidney CT and develop a framework to address it. The task takes a single 3D kidney CT volume and outputs a variable number of lesions each described by four clinical attributes.
We construct a dataset of 2,619 CT volumes from 788 patients at a single academic medical center. Each volume is paired with a radiology report. Lesion-centric attribute labels (type, size, enhancement, attenuation) are extracted from the reports. A segmentation model additionally produces masks for the kidney, cyst, and solid lesions, along with a cropped bounding box around the kidney region. Manual verification is integrated into the annotation pipeline to reduce label noise. The resulting labels are organized at three granularity levels: side-level abnormality (L1), side-level lesion typing and size (L2), and per-lesion attributes (L3). The CT inputs are correspondingly organized at several levels: whole CT, cropped CT, and segmentation-mask channels. This hierarchical structure of both inputs and labels supports controlled ablations across input and label granularities.
The framework pairs pretrained 3D encoders and segmentation-mask inputs with LesionDETR, a transformer-based set-prediction architecture. This head is trained with a hierarchical supervision loss that semantically links per-lesion predictions with coarser side-level objectives (Figure~\ref{fig:pipeline}).

Evaluating per-lesion performance is non-trivial because no standard metric handles a variable number of predictions against a variable number of ground-truth lesions. We adapt COCO's detection protocol to our setting, replacing the IoU geometric gate with a size-distance tolerance and validating the metric on synthetic test cases (Section~\ref{sec:metrics}).
We conducted extensive experiments spanning six encoder initializations, varying input and prediction granularities, and external generalization on the KiTS23 dataset. We hope this work provides a new framework for lesion detection in 3D CT and supports downstream tasks such as medical report generation.
Our contributions are:
\begin{enumerate}
    \item \textbf{Lesion-centric task and evaluation.} We formulate kidney CT characterization as a lesion-centric set-prediction task: per kidney, the model predicts a variable number of lesions, each described by four clinical attributes. To evaluate this task under variable cardinality, we adapt the COCO detection protocol with size-distance tolerance and per-class Average Precision.

    \item \textbf{Dataset with hierarchical inputs and labels.} We construct a high-quality dataset with lesion-centric attribute labels, segmentation masks, and a cropped bounding box around the kidney. To ensure clinical relevance, mass and tumor attributes are merged into a single \emph{solid} class based on clinical terminology overlap. Both inputs and labels are organized at multiple granularities, enabling systematic study of input-label granularity interactions. We additionally use the KiTS23 cohort (489 cases) as an independent external validation set.

    \item \textbf{Per-lesion detection model (LesionDETR).} We propose LesionDETR, a transformer-based detection head utilizing learnable queries for variable-count lesion prediction, with size-based Hungarian matching in place of IoU. A hierarchical loss couples per-lesion outputs with side-level objectives, enabling supervision at multiple granularity levels from a single model.

    \item \textbf{Design principles.} Controlled experiments identify four dominant design factors: (i) a segmentation mask as an input channel helps despite its upstream noise; (ii) abdominal pretraining (SuPreM) transfers, while generic large-corpus pretraining does not; (iii) hierarchical supervision lets one model serve all three granularities, but aggregated outputs trail direct supervision at each target level; (iv) a trade-off between side-level aggregation and per-lesion detection emerges across head architectures.
\end{enumerate}

\section{Related Work}
\label{sec:related}

\subsection{Kidney CT analysis and classification}

Prior deep learning work on kidney CT classification operates at the patient or organ level. 2D approaches have addressed renal mass characterization from cropped lesion regions or individual slices, including Bosniak cyst stratification~\cite{xi2020stratification}, renal cancer subtype classification~\cite{han2019classification,klontzas2024cnn}, and cystic lesion malignancy prediction~\cite{uhlig2022deep}. In 3D, Zhou et al.~\cite{zhou2021endtoend} developed an end-to-end 3D pipeline for tumor subtype differentiation from multi-phase CT, and LACPANet~\cite{lacpanet2024} introduced cross-phase attention for the same task. These methods produce a single prediction per case and do not model multiple lesions or per-lesion attributes.

\subsection{Segmentation and lesion localization in 3D CT}

Kidney segmentation from 3D CT has reached high accuracy. nnU-Net~\cite{isensee2021nnunet} exceeds Dice 0.97 for kidney parenchyma and 0.83--0.85 for tumors on the KiTS benchmark~\cite{myronenko2023kits}. TotalSegmentator~\cite{wasserthal2023totalseg} provides whole-body segmentation including kidneys. Segmentation outputs have also been used to guide downstream tasks~\cite{yan2019mulan,zlocha2019improving,oktay2018attention,kamnitsas2017efficient}, with common strategies including mask-as-channel, mask-guided pooling, and mask-guided attention. In chest CT, organ-level features outperform global representations~\cite{ctgraph2025,baharoon2025design}. These methods focus on spatial boundaries and do not capture structured clinical attributes per lesion.

\subsection{Representation learning and report-based supervision}

Transfer learning is widely used in medical image analysis~\cite{tajbakhsh2016cnn,raghu2019transfusion}. For 3D CT, several pretrained encoders are available: SuPreM~\cite{li2024suprem} uses supervised pretraining on 9.2K abdominal CTs for 25-class segmentation, VoCo~\cite{wu2024voco} employs self-supervised volume contrastive learning on 160K CTs, CT-CLIP~\cite{hamamci2024ctclip} provides vision-language pretraining on chest CT-report pairs, and Models Genesis~\cite{zhou2021models} showed that self-supervised 3D pretraining can outperform ImageNet transfer. These encoders have shown strong performance on segmentation benchmarks~\cite{huang2023stunet,blankemeier2026merlin,bai2024m3d}. On the supervision side, clinical reports can serve as a source of structured labels. CheXpert~\cite{irvin2019chexpert} introduced rule-based label extraction from radiology reports, and Adams et al.~\cite{adams2023leveraging} showed LLM-based structured reporting. RadGPT~\cite{bassi2025radgpt} generates tumor descriptions through template filling from segmentation masks, and Liang et al.~\cite{liang2025kidney} proposed a two-stage framework for kidney CT report generation from 2D slices. These approaches treat report generation as an end-to-end task, using the full report as the supervision signal. A lesion-centric approach that produces structured per-lesion predictions could complement these methods by providing verified intermediate representations, enabling more controllable and interpretable report generation.

\subsection{Multi-instance and set prediction methods}

Predicting a variable number of structured outputs per input is a recurring problem in medical imaging. Multiple-instance learning (MIL) provides a framework for reasoning about multiple instances within a sample, with aggregation strategies for deriving sample-level predictions from instance-level outputs~\cite{zhao2023multitask}. DETR~\cite{carion2020detr} introduced set prediction with learnable queries and bipartite matching, offering a way to handle variable-count outputs without predefined ordering, and has been applied to medical lesion detection~\cite{li2025detr_medical}. Beyond methods, the COCO benchmark~\cite{lin2014coco} established standard evaluation conventions for set-prediction tasks (per-class Average Precision computed via greedy-by-confidence matching with an IoU-based geometric gate), which we adapt to our setting.

\section{Materials and Methods}
\label{sec:methods}

\subsection{Study cohort}
\label{sec:cohort}
We retrospectively collected renal CT data from the UF Health Integrated Data Repository (IDR)\footnote{\url{https://idr.ufhealth.org/}}, selecting studies associated with renal-related CPT codes (74160, 74170, 74175, 74177, 74178) for CT scans that include the abdomen, dated between December 2, 2011 and August 24, 2024. This initial query yielded 2,297 radiology reports from 896 patients. Of these, 391 reports had no available CT scan data and were excluded, resulting in 1,906 reports linked to 3,101 CT volumes from 794 patients. This study was approved by UF's Institutional Review Board (IRB202400720), and the requirement for written informed consent was waived. All procedures were conducted in accordance with the Declaration of Helsinki. 

The cohort (Table~\ref{tab:demographics}) consists predominantly of older adults (median age 65) with a near-balanced sex distribution and primarily White or Black/African American racial composition. Median kidney function is mildly to normally impaired, but \textasciitilde22\% of patients have CKD stage G3 or worse, higher than the general population and reflecting that renal CT is typically ordered for patients with known or suspected kidney pathology. Hypertension and diabetes are the dominant comorbidities. Train and validation sets show no significant distributional differences (all $p > 0.05$), supporting the validity of the patient-level split.

\begin{table}[!htbp]
\centering
\scriptsize
\setlength{\tabcolsep}{5pt}
\renewcommand{\arraystretch}{1.05}
\caption{Demographics and clinical characteristics of the kidney CT cohort. Continuous variables are reported as median [IQR]; categorical variables as n (\%). \textit{p}-values for train/validation balance.}
\label{tab:demographics}
\begin{threeparttable}
\begin{tabular}{lllll}
\toprule
Variable & Train (n=632) & Validation (n=156) & Total (n=788) & \textit{p} \\
\midrule
Age (years)\tnote{*} & 65.0 [59.0--70.0] & 65.0 [60.0--70.0] & 65.0 [59.0--70.0] & 0.813 \\
BMI (kg/m\textsuperscript{2}) & 27.3 [23.4--32.4] & 28.2 [23.4--33.0] & 27.6 [23.4--32.5] & 0.299 \\
Creatinine (mg/dL) & 0.9 [0.8--1.2] & 1.0 [0.8--1.2] & 1.0 [0.8--1.2] & 0.561 \\
eGFR (mL/min/1.73 m\textsuperscript{2}) & 78.1 [60.5--95.2] & 74.5 [59.3--94.4] & 77.4 [59.8--95.1] & 0.362 \\
Sex &  &  &  & 0.279 \\
\quad Male & 357 (56.5) & 80 (51.3) & 437 (55.5) &  \\
\quad Female & 275 (43.5) & 76 (48.7) & 351 (44.5) &  \\
Race &  &  &  & 0.110 \\
\quad White & 463 (73.3) & 106 (67.9) & 569 (72.2) &  \\
\quad Black/AA & 143 (22.6) & 49 (31.4) & 192 (24.4) &  \\
\quad Asian & 5 (0.8) & 0 (0.0) & 5 (0.6) &  \\
\quad AI/AN & 3 (0.5) & 0 (0.0) & 3 (0.4) &  \\
\quad More than one race & 1 (0.2) & 0 (0.0) & 1 (0.1) &  \\
\quad Other & 17 (2.7) & 1 (0.6) & 18 (2.3) &  \\
Ethnicity &  &  &  & 0.680 \\
\quad Hispanic & 12 (1.9) & 2 (1.3) & 14 (1.8) &  \\
\quad Non-Hispanic & 618 (97.8) & 154 (98.7) & 772 (98.0) &  \\
\quad Unknown & 2 (0.3) & 0 (0.0) & 2 (0.3) &  \\
CKD stage (KDIGO eGFR) &  &  &  & 0.659 \\
\quad G1 & 186 (29.4) & 37 (23.7) & 223 (28.3) &  \\
\quad G2 & 229 (36.2) & 60 (38.5) & 289 (36.7) &  \\
\quad G3a & 75 (11.9) & 18 (11.5) & 93 (11.8) &  \\
\quad G3b & 36 (5.7) & 9 (5.8) & 45 (5.7) &  \\
\quad G4 & 13 (2.1) & 2 (1.3) & 15 (1.9) &  \\
\quad G5 & 14 (2.2) & 6 (3.8) & 20 (2.5) &  \\
\quad Unknown & 79 (12.5) & 24 (15.4) & 103 (13.1) &  \\
Smoking &  &  &  & 0.873 \\
\quad Current & 263 (41.6) & 65 (41.7) & 328 (41.6) &  \\
\quad Former & 362 (57.3) & 90 (57.7) & 452 (57.4) &  \\
\quad Never & 7 (1.1) & 1 (0.6) & 8 (1.0) &  \\
Hypertension & 477 (75.5) & 118 (75.6) & 595 (75.5) & 1.000 \\
Diabetes & 224 (35.4) & 63 (40.4) & 287 (36.4) & 0.291 \\
CKD diagnosis (ICD-10 N18.x) & 159 (25.2) & 46 (29.5) & 205 (26.0) & 0.316 \\
Renal cancer history & 17 (2.7) & 6 (3.8) & 23 (2.9) & 0.429 \\
Prior nephrectomy & 2 (0.3) & 0 (0.0) & 2 (0.3) & 1.000 \\
\bottomrule
\end{tabular}
\begin{tablenotes}[flushleft]
\footnotesize
\item[] AI/AN denotes American Indian or Alaska Native; CKD denotes chronic kidney disease. \textsuperscript{*}Age data missing for 19 patients due to incomplete OMOP visit linkage.
\end{tablenotes}
\end{threeparttable}
\end{table}

\subsection{Dataset processing}
\label{sec:dataproc}

\begin{figure}[!htbp]
    \centering
    \includegraphics[width=\textwidth]{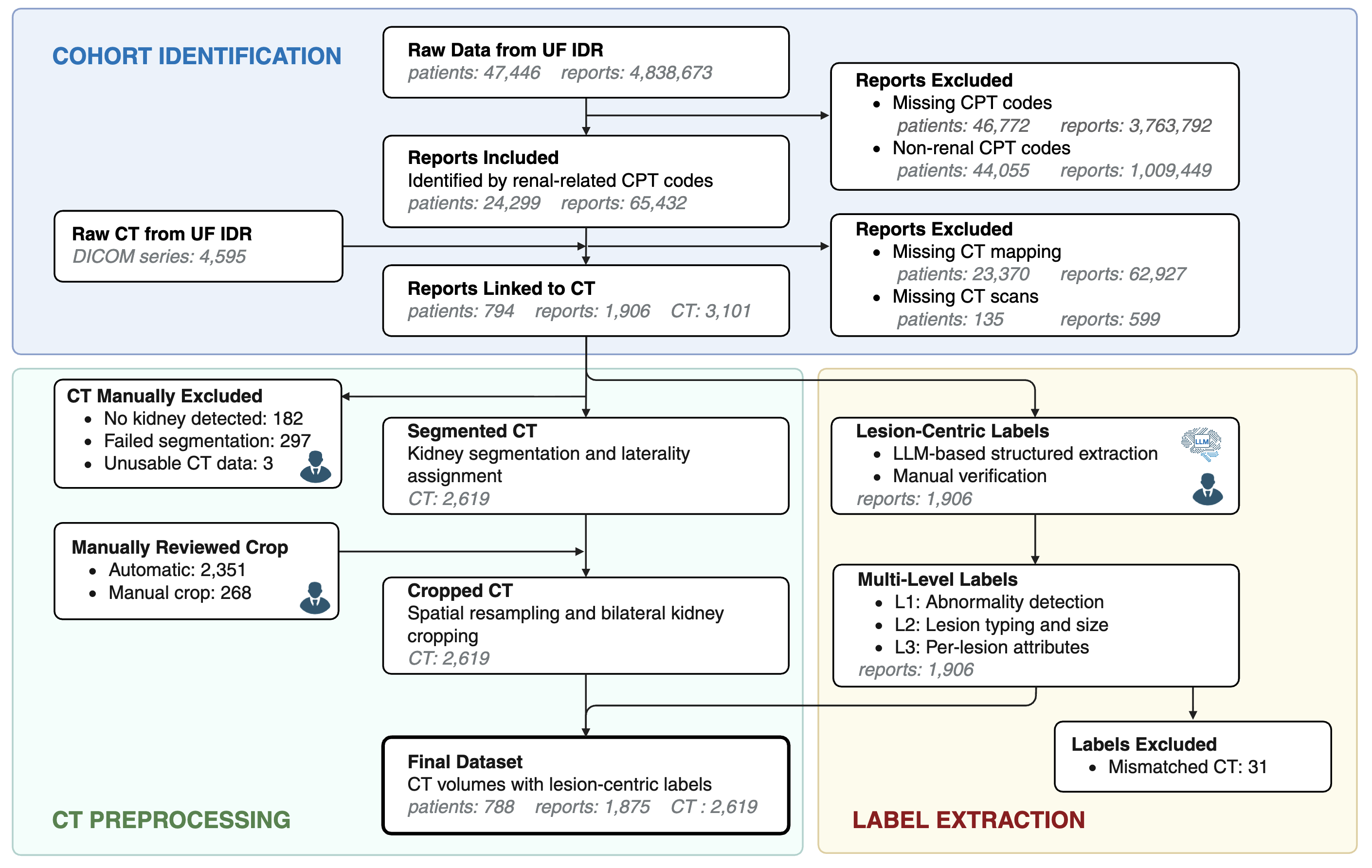}
    \caption{Data processing workflow. \textbf{(A) Cohort identification.} Renal CT studies selected from the UF Health IDR by CPT code. \textbf{(B) CT preprocessing.} Segmentation, resampling, cropping, and two manual review stages. \textbf{(C) Lesion-centric label extraction.} An LLM extracted per-lesion attributes from radiology reports, followed by two-annotator verification. The two pipelines converge into a dataset of 788 patients with multi-level annotations.}
    \label{fig:funnel}
\end{figure}

Following cohort identification, the data are processed through two parallel pipelines: a CT preprocessing pipeline that prepares the imaging volumes for model input, and a label extraction pipeline that derives lesion-centric labels from the radiology reports. Figure~\ref{fig:funnel} illustrates the complete workflow.

\subsubsection{CT preprocessing}
\label{sec:ctpreproc}

Each volume was segmented with an nnU-Net model~\cite{isensee2021nnunet} pretrained on the KiTS challenge data to produce multi-class masks separating kidney parenchyma, tumor, and cyst on each side. The volumes were then resampled to 1 mm isotropic spacing and cropped to the bilateral kidney region using the segmentation mask. Intensity values were windowed to the soft-tissue range $[-200, 400]$~HU. Full preprocessing parameters are provided in the supplementary material.

Two manual review stages were inserted to ensure data quality. The first stage, after segmentation, verified that every retained volume contained clearly visible kidney regions; volumes with no kidney detected, failed segmentation, or other unusable data were excluded. The second stage, after automatic cropping, verified the bounding box correctness; suboptimal crops were manually adjusted. After review, we are confident that every CT volume in the final dataset contains anatomically valid kidney regions. The kidney parenchyma masks are reliable by construction, while the tumor and cyst masks, which inherit the biases of the nnU-Net model trained on a different clinical population, are silver labels with known noise. A detailed cross-modal comparison between mask-derived findings and report-derived labels is provided in the supplementary material.
The final dataset consists of 2,619 volumes from 788 patients, split at the patient level into training (2,167 volumes, 632 patients) and validation (452 volumes, 156 patients) sets.

\subsubsection{Lesion-centric label extraction}
\label{sec:labels}

Lesion-centric labels were extracted from the radiology report of each clinical exam using a large language model (LLM) pipeline, similar to Adams et al.~\cite{adams2023leveraging} for structured radiology reporting. The renal portion of each report was passed to the LLM, which directly extracted per-lesion attributes (L3): type (cyst, mass, or tumor), size, enhancement pattern, and attenuation. Because mass and tumor are individually too rare for stable separate evaluation, we merge them into a single \emph{solid} category in consultation with clinicians; this is consistent with how the terms are used interchangeably in clinical practice. Note that the cyst and solid flags are not mutually exclusive: Bosniak III/IV complex cystic lesions legitimately carry both flags. A small fraction of lesions describe AML, lipoma, or infarct, which fall outside the cyst/solid schema and carry neither flag. Qualitative size descriptors in the reports (``small''/``subcentimeter'' and ``large'') are mapped to clinician-validated numeric values (0.8 cm and 3.0 cm, respectively), so that all lesions carry a continuous size for regression and per-lesion matching. Side-level labels at L2 (cystic or solid presence and maximum lesion size per kidney) and L1 (binary abnormality per kidney) were then computed deterministically from the L3 attributes.

Two annotators were involved in verification. The first annotator reviewed and corrected the LLM output for every report.  A second reviewer performed secondary verification of the corrected labels on 100 sampled reports. Agreement was high, with a macro-average Cohen's $\kappa$ of 0.969 for categorical attributes and an intraclass correlation coefficient (ICC) of 1.000 for lesion size. Most disagreements involved borderline cyst labels, particularly cases initially marked as probable cysts that were revised to definite cysts, while lesion size was unchanged during secondary review.

\subsubsection{Dataset statistics}
\label{sec:stats_data}

Table~\ref{tab:label_dist} summarizes the label distribution of the UF Health dataset at the cohort and side level. The L3 layer contains 2,375 individual lesions in total, each annotated with cyst, mass, tumor, size, enhancement, and attenuation attributes. After consolidating mass and tumor into a single solid category, cysts dominate the lesion population while solid lesions remain rare. The per-lesion count is small in practice: most kidney sides contain zero or one lesion, and fewer than 1\% of exams report more than three lesions on a single side. This observation justifies the fixed three-slot design used in the per-lesion prediction head (Section~\ref{sec:multigranularity}). Full L3 attribute distributions are reported in the supplementary material.

\begin{table}[!htbp]
\centering
\scriptsize
\setlength{\tabcolsep}{6pt}
\renewcommand{\arraystretch}{1.1}
\caption{Label distribution of the UF Health kidney CT dataset. Labels are reported at the exam level (deduplicated); train and validation sets are split at the patient level. Solid is the union of Mass and Tumor flags.}
\label{tab:label_dist}
\begin{tabular}{lrrr}
\toprule
& \textbf{Train} & \textbf{Valid} & \textbf{Total} \\
\midrule
Patients & 632 & 156 & 788 \\
CT volumes & 2,167 & 452 & 2,619 \\
Reports & 1,539 & 336 & 1,875 \\
\midrule
\multicolumn{4}{l}{\textit{L1 -- Side-level abnormality}} \\
\quad Left abnormal & 839 & 194 & 1,033 \\
\quad Right abnormal & 776 & 201 & 977 \\
\midrule
\multicolumn{4}{l}{\textit{L2 -- Side-level lesion type}} \\
\quad Left cyst & 816 & 189 & 1,005 \\
\quad Right cyst & 753 & 199 & 952 \\
\quad Left solid & 58 & 9 & 67 \\
\quad Right solid & 40 & 4 & 44 \\
\midrule
\multicolumn{4}{l}{\textit{L3 -- Per-lesion}} \\
\quad Total lesions & 1,907 & 468 & 2,375 \\
\quad with Cyst flag & 1,827 & 454 & 2,281 \\
\quad with Mass flag & 99 & 13 & 112 \\
\quad with Tumor flag & 43 & 4 & 47 \\
\quad with Solid flag (Mass $\cup$ Tumor) & 101 & 13 & 114 \\
\midrule
\multicolumn{4}{l}{\textit{L3 -- Per-side lesion count distribution}} \\
\quad 0 lesions & 1463 (47.5\%) & 277 (41.2\%) & 1740 (46.4\%) \\
\quad 1 lesion & 1346 (43.7\%) & 327 (48.7\%) & 1673 (44.6\%) \\
\quad 2 lesions & 249 (8.1\%) & 64 (9.5\%) & 313 (8.3\%) \\
\quad $\geq$3 lesions & 20 (0.6\%) & 4 (0.6\%) & 24 (0.6\%) \\
\midrule
\multicolumn{4}{l}{\textit{L3 -- Lesion size}} \\
\quad Median size (cm) [IQR] & 1.1 [0.8--2.4] & 1.3 [0.8--3.0] & 1.2 [0.8--2.5] \\
\midrule
\multicolumn{4}{l}{\textit{L3 -- Enhancement}} \\
\quad Enhancing & 63 (3.3) & 10 (2.1) & 73 (3.1) \\
\quad Non-enhancing & 85 (4.5) & 19 (4.1) & 104 (4.4) \\
\quad Unknown / not reported & 1,759 (92.2) & 439 (93.8) & 2,198 (92.5) \\
\midrule
\multicolumn{4}{l}{\textit{L3 -- Attenuation}} \\
\quad Hyperdense & 97 (5.1) & 16 (3.4) & 113 (4.8) \\
\quad Isodense & 17 (0.9) & 1 (0.2) & 18 (0.8) \\
\quad Hypodense & 597 (31.3) & 133 (28.4) & 730 (30.7) \\
\quad Unknown / not reported & 1,196 (62.7) & 318 (67.9) & 1,514 (63.7) \\
\bottomrule
\end{tabular}
\end{table}

\subsection{Model architecture}
\label{sec:model}

Figure~\ref{fig:pipeline} gives an overview of the framework. A 3D CT volume, optionally combined with a kidney segmentation mask, is encoded by a 3D CT encoder and decoded by multi-level prediction heads, with a hierarchical loss linking the per-lesion outputs to side-level predictions. The following subsections describe each component in detail.

\subsubsection{Input representation design}
\label{sec:input}

We compare four input representations along two dimensions: whether the volume is cropped to the kidney region, and whether segmentation information is provided as an extra input channel. \textit{Whole CT} uses the full single-channel CT volume. \textit{Cropped CT} bilaterally crops around the kidney region using the segmentation bounding box. \textit{Cropped CT + Kidney mask} additionally concatenates a binary kidney mask as a second channel, while \textit{Cropped CT + Full mask} concatenates the original multi-class segmentation mask. The first two configurations isolate the effect of spatial localization; the latter two add segmentation as an explicit prior, with the kidney mask preserving only the kidney boundary and the full mask retaining the lesion-class labels.

\subsubsection{Encoder configurations}
\label{sec:encoders}

We compare encoder configurations spanning four pretraining strategies.

\begin{itemize}
    \item \textbf{SuPreM}~\cite{li2024suprem} (supervised domain): SwinUNETR~\cite{tang2022swinunetr} (62M) pretrained with multi-organ segmentation on 9.2K abdominal CTs.
    \item \textbf{VoCo}~\cite{wu2024voco} (self-supervised domain): The same SwinUNETR architecture, pretrained with volume contrastive learning on 160K CTs.
    \item \textbf{CTViT}~\cite{hamamci2024ctclip} (cross-modal): The vision encoder of CT-CLIP, pretrained with contrastive image-text alignment on CT-RATE.
    \item \textbf{From-scratch baselines}: SwinUNETR, CTViT, and a 3D ResNet-18 (33M), all randomly initialized.
\end{itemize}

All encoders are followed by global average pooling to produce a fixed-length feature vector.

\subsubsection{Per-lesion detection head}
\label{sec:detection_head}

The L3 head, \textbf{LesionDETR}, must produce a variable number of per-lesion predictions per kidney; we adopt a DETR-style design built on learnable queries and Hungarian matching.

As shown in Figure~\ref{fig:l3_head}A, the head uses $N$ learnable query vectors per kidney side ($N=3$, giving 6 queries total). Each query is refined by multi-head cross-attention to the encoder's spatial feature map (rather than to a single pooled vector), giving each query access to spatially distributed evidence. A small MLP then maps each refined query to per-slot predictions: existence, two type flags (cyst, solid), size, and two categorical attributes (enhancement, attenuation).

At training time, predicted slots are paired with ground-truth lesions via the Hungarian algorithm. We replace DETR's IoU-based geometric cost with an absolute size-distance term, augmented by classification losses on the cyst and solid attributes. Predictions not matched to any ground truth are designated as ``no-object'' and supervised toward existence $=0$ (weight $0.1$, following the DETR default). Figure~\ref{fig:l3_head}B illustrates a matching example with two ground-truth lesions and four prediction slots.

DETR-style set-prediction matches two properties of this task: the number of lesions per kidney is variable, and the learnable queries adapt to lesion-specific spatial evidence at training time. Simpler alternatives are included in Section~\ref{sec:results_detection} as ablations.

\subsubsection{Multi-granularity outputs and hierarchical supervision}
\label{sec:multigranularity}

The prediction head produces outputs at all three label granularities. For L1 and L2, simple fully-connected layers attached to the pooled encoder feature produce two binary outputs for L1 and four binary plus two regression outputs for L2. For L3, the DETR-style head described above produces six attributes per slot.

\begin{figure}[!htbp]
    \centering
    \includegraphics[width=\textwidth]{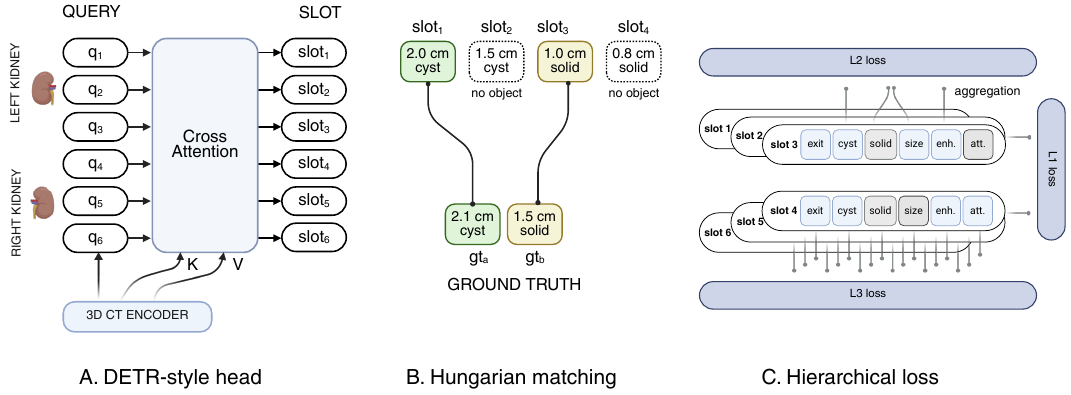}
    \caption{Per-lesion detection head and hierarchical supervision. \textbf{(A)} DETR-style head: learnable query vectors (three per kidney side) refined by cross-attention to the encoder's feature map and decoded into per-slot predictions. \textbf{(B)} Hungarian matching: predicted slots are paired with ground-truth lesions using a size-distance cost; unmatched slots are designated ``no object''. \textbf{(C)} Hierarchical supervision: each slot is supervised at L3, while L1 and L2 losses supervise side-level probabilities aggregated from per-slot outputs.}
    \label{fig:l3_head}
\end{figure}

A model trained only at L3 receives no direct supervision for the coarser L1 and L2 outputs, even though the slot predictions implicitly contain that information. We add a \emph{hierarchical supervision} loss that aggregates per-slot probabilities into side-level probabilities and supervises them against the L1/L2 ground truth (Figure~\ref{fig:l3_head}C). The aggregation uses the complement product rule, the differentiable analogue of a logical OR over the slots: a side is abnormal if at least one of its slots contains a lesion. Letting $p_k$ denote the predicted existence probability for slot $k$ on a given side,
\begin{equation}
    P(\text{side abnormal}) = 1 - \prod_{k=1}^{3}(1 - p_k).
\end{equation}
The same rule applies to side-level cyst and solid predictions by replacing $p_k$ with the corresponding per-slot attribute probability.

The total loss combines the L3 set-prediction loss with the hierarchical side-level loss:
\begin{equation}
    \mathcal{L} = \mathcal{L}_{\text{L3}} + \lambda \, \mathcal{L}_{\text{hier}},
\end{equation}
where $\mathcal{L}_{\text{L3}}$ is the per-slot set-prediction loss from the Hungarian matching and $\mathcal{L}_{\text{hier}}$ is binary cross-entropy on the aggregated side-level probabilities. We ablate $\lambda$ and which side-level losses to apply (L1 only vs.\ L1+L2) in Section~\ref{sec:results_hier}. A single L3-trained model can then be evaluated at all three granularity levels without retraining.

\subsection{Experimental setup}
\label{sec:expsetup}

Models were trained for 30 epochs without early stopping using AdamW with a cosine annealing schedule and mixed-precision (fp16) autocast, on NVIDIA B200 GPUs. The encoder is trained at $0.1\times$ the head learning rate, which we found necessary to preserve pretrained features under DETR-style set-prediction training. We applied intensity-only augmentation and avoided spatial augmentations such as flipping or rotation, since these would alter the laterality and spatial attributes being predicted. Key experiments are repeated with three random seeds, and bootstrap confidence intervals (1,000 resamples) are reported for AUC differences. Full hyperparameters are provided in the supplementary material.

\subsection{Evaluation metrics}
\label{sec:metrics}

We evaluate models at all three granularity levels, with a different metric family for each.

\paragraph{Side-level metrics.} For binary attributes we report AUC per kidney. Lesion size at L2 is evaluated with mean absolute error (MAE) in centimeters against the largest lesion per kidney. These metrics directly evaluate L1- and L2-trained models; an L3-trained model's per-slot predictions can additionally be aggregated at the per-kidney level via
\begin{equation}
  P_{\text{kidney}}^{(a)} \;=\; 1 - \prod_{i} \bigl(1 - p_{a,i}\bigr),
  \label{eq:per_kidney_agg}
\end{equation}
where $p_{a,i}$ is the per-slot probability of attribute $a$ at slot $i$. This lets us ask whether hierarchical supervision helps side-level prediction.

\paragraph{Per-lesion metrics.} A per-lesion metric must measure whether individual lesions are correctly identified and characterized. This decomposes into two problems: (i) matching a variable number of predictions against a variable number of ground-truth lesions per kidney, and (ii) scoring how accurately each matched prediction characterizes its lesion. Inspired by multi-instance matching tasks such as COCO~\cite{lin2014coco}, ranking is driven by class confidence and match acceptance by size distance (substituting for COCO's IoU threshold).

On each kidney, predictions are greedily matched to ground-truth lesions in descending order of confidence. A match is accepted only if the size distance between the prediction and the ground truth is within a tolerance $\tau$. Unmatched predictions count as false positives; unmatched ground-truth lesions count as false negatives.

Given a match, the score quantifies how confidently the model predicts the correct class at the matched location. Because AP ranks predictions by this score, it must reflect classifier quality rather than match quality.

Let slot $i$ on a kidney emit existence logit $e_i$, per-class type logits $t_{c,i}$, and predicted size $\hat s_i$. The class-$c$ confidence is
\begin{equation}
  \mathrm{conf}_c(i) \;=\; \sigma(e_i)\,\sigma(t_{c,i}),
  \label{eq:slot_conf}
\end{equation}
i.e.\ the posterior of a class-$c$ lesion at slot $i$ under independent existence and type decisions. Predictions are ranked by Eq.~\eqref{eq:slot_conf} and greedily matched to class-$c$ ground-truth lesions $j$ with size $s_j$; a match is accepted only if $|\hat s_i - s_j| \le \tau$ and $j$ is not already claimed. Let $p_c(r)$ denote the resulting precision at recall $r$ for class $c$; we report the 11-point Pascal interpolation
\begin{equation}
  \mathrm{AP}_c \;=\; \tfrac{1}{11}\!\!\sum_{r \in \{0,0.1,\ldots,1.0\}} \max_{\tilde r \ge r} p_c(\tilde r),
  \label{eq:ap11}
\end{equation}
and take mAP as the mean of $\{\mathrm{AP}_c\}$ over classes with $n_{\text{pos}}>0$ (classes with no positives are undefined and excluded, following COCO). In our setting, $c \in \{\text{cyst}, \text{solid}\}$. We sweep size tolerances $\tau \in \{0.5, 1.0, 2.0\}$ cm, analogous to the AP\textsubscript{50}, AP\textsubscript{75}, and AP threshold sweep used in COCO. A synthetic validation of this protocol (golden test cases with analytically known AP and four sensitivity sweeps) is given in Appendix~\ref{app:eval_validation}.

\section{Results}
\label{sec:results}

\subsection{Side-level results}
\label{sec:results_corematrix}

We trained 12 models, one for each pairing of four input representations with three training granularities (L1, L2, L3). For L3-trained models, side-level L1 and L2 metrics are obtained by hierarchical aggregation rather than direct supervision. Table~\ref{tab:corematrix} summarizes the side-level results, from which two patterns are visible.

Within each training block, adding spatial information to the input helps: cropping to the kidney region gives a moderate gain, and concatenating a segmentation mask gives a much larger one. At L1 and L2 the two mask variants perform comparably on side-level metrics, with the full multi-class mask marginally ahead on bilateral L2 cyst AUC and size MAE. Under L3 supervision, both cropped and mask inputs improve side-level AUCs after hierarchical aggregation.

\begin{table*}[!htbp]
\centering
\scriptsize
\setlength{\tabcolsep}{3.5pt}
\renewcommand{\arraystretch}{1.05}
\caption{Side-level and per-lesion metrics across four input representations and three training granularities. L1/L2 columns report AUC; L2 Size reports MAE (cm); L3 columns report per-class and mean Average Precision at size tolerance $\tau = 1\,$cm.}
\label{tab:corematrix}
\begin{tabular}{l ccccc ccc}
\toprule
& \multicolumn{2}{c}{\textbf{Evaluated as L1}} & \multicolumn{3}{c}{\textbf{Evaluated as L2}} & \multicolumn{3}{c}{\textbf{Evaluated as L3}} \\
\cmidrule(lr){2-3} \cmidrule(lr){4-6} \cmidrule(lr){7-9}
\textbf{CT Input} & Left abn.\,$\uparrow$ & Right abn.\,$\uparrow$ & Left cyst\,$\uparrow$ & Right cyst\,$\uparrow$ & Size\,$\downarrow$ & AP$_{\text{cyst}}$\,$\uparrow$ & AP$_{\text{solid}}$\,$\uparrow$ & mAP\,$\uparrow$ \\
\midrule
\multicolumn{9}{l}{\textit{Trained at L1 }} \\
Whole CT      & 0.647 & 0.612 & --- & --- & --- & --- & --- & --- \\
Cropped CT    & 0.727 & 0.697 & --- & --- & --- & --- & --- & --- \\
+ Kidney mask & \textbf{0.873} & 0.802 & --- & --- & --- & --- & --- & --- \\
+ Full mask   & 0.852 & \textbf{0.828} & --- & --- & --- & --- & --- & --- \\
\midrule
\multicolumn{9}{l}{\textit{Trained at L2 }} \\
Whole CT      & 0.661 & 0.595 & 0.664 & 0.599 & 1.111 & --- & --- & --- \\
Cropped CT    & 0.687 & 0.679 & 0.692 & 0.681 & 0.953 & --- & --- & --- \\
+ Kidney mask & 0.843 & 0.795 & 0.843 & 0.797 & 0.790 & --- & --- & --- \\
+ Full mask   & \textbf{0.860} & \textbf{0.801} & \textbf{0.858} & \textbf{0.801} & \textbf{0.770} & --- & --- & --- \\
\midrule
\multicolumn{9}{l}{\textit{Trained at L3}} \\
Whole CT      & 0.651 & 0.633 & 0.659 & 0.636 & 1.439 & 0.150 & \textbf{0.053} & \textbf{0.102} \\
Cropped CT    & \textbf{0.817} & 0.781 & \textbf{0.826} & 0.778 & 1.460 & 0.145 & 0.020 & 0.082 \\
+ Kidney mask & 0.777 & 0.726 & 0.788 & 0.723 & 1.392 & \textbf{0.165} & 0.015 & 0.090 \\
+ Full mask   & 0.801 & \textbf{0.804} & 0.807 & \textbf{0.803} & \textbf{1.257} & 0.155 & 0.037 & 0.096 \\
\bottomrule
\end{tabular}
\end{table*}

Training at a finer granularity supports richer outputs, but aggregating these back to coarser levels does not fully recover the accuracy of direct supervision. Evaluated as L1, L1-trained models outperform L3-aggregated predictions by roughly $0.04$ bilateral AUC. Evaluated as L2, L2-trained models outperform L3 aggregation by $0.02$--$0.05$ AUC.

Overall, providing additional spatial information consistently helps despite the noise inherited from the upstream segmentation: even the full multi-class mask, which carries more channel-specific noise than the binary kidney mask, yields measurable gains. However, the appropriate level of spatial prior depends on the training granularity: the full mask is preferable at side-level training (L1/L2) where its richer class channels outweigh their noise, whereas at L3 the slot-level supervision already extracts lesion-class signal from CT intensity and a simpler cropped or binary-kidney input is sufficient. The input choice should therefore be calibrated to the training granularity.
\FloatBarrier

\subsection{Per-lesion detection results}
\label{sec:results_detection}

This section compares the detection-oriented refinements by head architecture. We evaluate four heads under a common training pipeline, reporting each head's best-performing configuration for the architecture comparison. All runs use cropped-CT input, SuPreM encoder, layered learning rate ($0.1\times$ on encoder), and three random seeds. We report mean $\pm$ standard deviation on the UF validation set.

The four heads differ in how slot identity and matching are assigned. \textbf{LesionDETR} uses $N=3$ learnable query vectors per kidney side, cross-attends to the encoder's spatial feature map, and matches predictions to ground-truth lesions through Hungarian assignment with a size-distance cost. The \textbf{count-conditioned head} first predicts the number of lesions per side, then uses fixed slot positional embeddings to produce per-slot attributes, also matched through Hungarian assignment. The \textbf{flatten-and-sort head} sorts ground-truth lesions by size and assigns them to fixed positional slots, with predictions supervised against the sorted ground truth without learned matching. The \textbf{summary head} produces a single per-side output by aggregating the encoder feature with no slot mechanism, and is a lower-bound baseline.

Table~\ref{tab:detection_ablation} reveals a three-way specialization across heads. LesionDETR wins every side-level metric (L1 abnormality AUC $0.799$) with the lowest seed variance, establishing it as the side-level specialist. The count-conditioned head flips this: it gives up side-level AUC to reach the highest per-lesion mAP ($0.190$), becoming the per-lesion specialist on common lesions. The flatten-and-sort head is the surprise: its non-learnable positional design, which sorts slots by predicted size, is the only configuration that clears the rare-class (solid) noise floor (AP$_{\text{solid}}$ $0.109$), suggesting that size-based ordering carries genuine signal for rare lesions. The summary head has no slot mechanism and trails all three slot-based alternatives on every metric, confirming that per-slot prediction is the lever.

\begin{table}[!htbp]
\centering
\scriptsize
\setlength{\tabcolsep}{5pt}
\renewcommand{\arraystretch}{1.15}
\caption{Head architecture comparison for per-lesion detection. Mean $\pm$ std across three seeds. L1 and L2 columns report bilateral AUC; AP columns report per-class average precision at size tolerance $\tau = 1$ cm.}
\label{tab:detection_ablation}
\begin{tabular}{l cccc c}
\toprule
\textbf{Architecture} & L1~abn$\uparrow$ & L2~cyst$\uparrow$ & L2~solid$\uparrow$ & AP$_{\text{cyst}}\uparrow$ & AP$_{\text{solid}}\uparrow$ \\
\midrule
LesionDETR   & $\mathbf{0.799}${\scriptsize$\pm 0.009$} & $\mathbf{0.802}${\scriptsize$\pm 0.007$} & $0.698${\scriptsize$\pm 0.042$} & $0.145${\scriptsize$\pm 0.013$} & $0.020${\scriptsize$\pm 0.006$} \\
Count-cond   & $0.690${\scriptsize$\pm 0.035$} & $0.595${\scriptsize$\pm 0.031$} & $0.498${\scriptsize$\pm 0.052$} & $\mathbf{0.369}${\scriptsize$\pm 0.074$} & $0.012${\scriptsize$\pm 0.003$} \\
Flatten-sort & $0.638${\scriptsize$\pm 0.020$} & $0.542${\scriptsize$\pm 0.010$} & $\mathbf{0.728}${\scriptsize$\pm 0.046$} & $0.187${\scriptsize$\pm 0.035$} & $\mathbf{0.109}${\scriptsize$\pm 0.013$} \\
Summary      & $0.639${\scriptsize$\pm 0.028$} & $0.566${\scriptsize$\pm 0.062$} & $0.676${\scriptsize$\pm 0.033$} & $0.072${\scriptsize$\pm 0.072$} & $0.076${\scriptsize$\pm 0.040$} \\
\bottomrule
\end{tabular}
\end{table}
\FloatBarrier

\subsection{Encoder comparison}
\label{sec:results_encoder}

We compare the six encoder configurations, all trained at L3 supervision with a kidney mask input, under identical optimization settings.
Two patterns stand out. First, pretraining helps only when task-aligned: SuPreM leads every other encoder on both mean bilateral AUC and size MAE, with a roughly 0.11 AUC gap over SwinUNETR trained from scratch. Second, abdominal-domain pretraining beats larger but generic corpora: VoCo is indistinguishable from random initialization, and CTViT offers only small gains. All remaining experiments therefore use SuPreM.

\begin{table}[!htbp]
\centering
\scriptsize
\setlength{\tabcolsep}{6pt}
\renewcommand{\arraystretch}{1.1}
\caption{Encoder comparison on L3-trained LesionDETR models with kidney mask input and hierarchical $L_1L_2$ supervision. Bilateral AUC averages left- and right-side AUCs; \textit{Mean} is the average of \textit{Abn} and \textit{Cyst}. \textit{Size} is MAE of the maximum-lesion prediction (cm). All values are mean $\pm$ std across three random seeds.}
\label{tab:encoder}
\begin{tabular}{l ccc c}
\toprule
\textbf{Encoder} & \multicolumn{3}{c}{\textbf{Bilateral AUC}\,$\uparrow$} & \textbf{Size (cm)}\,$\downarrow$ \\
\cmidrule(lr){2-4}
 & Abn & Cyst & Mean & \\
\midrule
\textbf{SuPreM}       & $\mathbf{0.768 \pm 0.027}$ & $\mathbf{0.773 \pm 0.027}$ & $\mathbf{0.771}$ & $\mathbf{1.27 \pm 0.22}$ \\
VoCo                  & $0.532 \pm 0.011$         & $0.528 \pm 0.012$         & $0.530$                  & $1.50 \pm 0.13$         \\
SwinUNETR scratch     & $0.541 \pm 0.055$         & $0.560 \pm 0.036$         & $0.551$                  & $1.45 \pm 0.02$         \\
\bottomrule
\end{tabular}
\end{table}

A layer-wise linear probing analysis of frozen encoder features (Appendix~\ref{app:probing}) localizes SuPreM's advantage to the mid-to-deep stages and confirms that a substantial fraction of its transferable signal is unlocked only through task-specific fine-tuning.
\FloatBarrier

\subsection{Hierarchical supervision ablation}
\label{sec:results_hier}

Table~\ref{tab:hier_ablation} compares three hierarchical supervision modes (no hier, L1 only, L1 + L2) on the two detection heads, with cropped CT input and three seeds per configuration. Hierarchical supervision substantially improves side-level AUCs for both heads: going from no hier to L1 + L2 raises L1 abnormality AUC by $0.135$ for LesionDETR ($0.642 \to 0.777$) and $0.152$ for the count-conditioned head ($0.584 \to 0.736$), with comparable gains on L2 cyst AUC. The per-slot aggregation therefore provides a useful gradient signal for side-level objectives even though those objectives are not supervised per slot.

These side-level gains come at a cost to per-lesion detection. For both heads, per-lesion mAP peaks at the L1-only mode and decreases once L2 supervision is added: LesionDETR mAP follows $0.072 \to 0.127 \to 0.093$ across the three modes, and the count-conditioned head follows $0.124 \to 0.171 \to 0.128$. AP$_{\text{cyst}}$ shows the same pattern, with both heads peaking under L1 only. L2 supervision therefore pushes the model toward side-level aggregation at the expense of per-slot discriminability, producing the same trade-off observed in the head architecture comparison (Table~\ref{tab:detection_ablation}). L2 solid is non-monotonic for both heads; given its small validation set (13 left-side and 7 right-side positives), we read this as high seed variance rather than a systematic effect.

\begin{table}[!htbp]
\centering
\scriptsize
\setlength{\tabcolsep}{5pt}
\renewcommand{\arraystretch}{1.15}
\caption{Hierarchical supervision ablation on two detection heads. Mean $\pm$ std across three seeds. L1 and L2 columns report bilateral AUC; AP columns report per-class average precision at $\tau = 1$ cm.}
\label{tab:hier_ablation}
\begin{tabular}{l cccccc}
\toprule
 & L1~abn$\uparrow$ & L2~cyst$\uparrow$ & L2~solid$\uparrow$ & AP$_{\text{cyst}}\uparrow$ & AP$_{\text{solid}}\uparrow$ & mAP$\uparrow$ \\
\midrule
\multicolumn{7}{l}{\textit{LesionDETR}} \\
no hier  & $0.642${\scriptsize$\pm 0.018$} & $0.647${\scriptsize$\pm 0.022$} & $0.655${\scriptsize$\pm 0.054$} & $0.117${\scriptsize$\pm 0.028$} & $0.027${\scriptsize$\pm 0.021$} & $0.072${\scriptsize$\pm 0.020$} \\
L1 only  & $0.691${\scriptsize$\pm 0.142$} & $0.676${\scriptsize$\pm 0.151$} & $0.585${\scriptsize$\pm 0.084$} & $0.190${\scriptsize$\pm 0.136$} & $0.064${\scriptsize$\pm 0.070$} & $0.127${\scriptsize$\pm 0.053$} \\
L1 + L2  & $\mathbf{0.777}${\scriptsize$\pm 0.022$} & $\mathbf{0.781}${\scriptsize$\pm 0.020$} & $0.644${\scriptsize$\pm 0.016$} & $0.162${\scriptsize$\pm 0.032$} & $0.024${\scriptsize$\pm 0.007$} & $0.093${\scriptsize$\pm 0.016$} \\
\midrule
\multicolumn{7}{l}{\textit{Count-cond}} \\
no hier  & $0.584${\scriptsize$\pm 0.035$} & $0.587${\scriptsize$\pm 0.026$} & $\mathbf{0.652}${\scriptsize$\pm 0.111$} & $0.234${\scriptsize$\pm 0.139$} & $0.014${\scriptsize$\pm 0.002$} & $0.124${\scriptsize$\pm 0.069$} \\
L1 only  & $0.708${\scriptsize$\pm 0.071$} & $0.679${\scriptsize$\pm 0.070$} & $0.483${\scriptsize$\pm 0.145$} & $\mathbf{0.320}${\scriptsize$\pm 0.152$} & $0.021${\scriptsize$\pm 0.016$} & $\mathbf{0.171}${\scriptsize$\pm 0.083$} \\
L1 + L2  & $0.736${\scriptsize$\pm 0.075$} & $0.740${\scriptsize$\pm 0.075$} & $0.643${\scriptsize$\pm 0.039$} & $0.190${\scriptsize$\pm 0.059$} & $\mathbf{0.066}${\scriptsize$\pm 0.083$} & $0.128${\scriptsize$\pm 0.013$} \\
\bottomrule
\end{tabular}
\end{table}
\FloatBarrier

\subsection{Data efficiency}
\label{sec:results_fewshot}

\begin{figure}[!htbp]
    \centering
    \includegraphics[width=\textwidth]{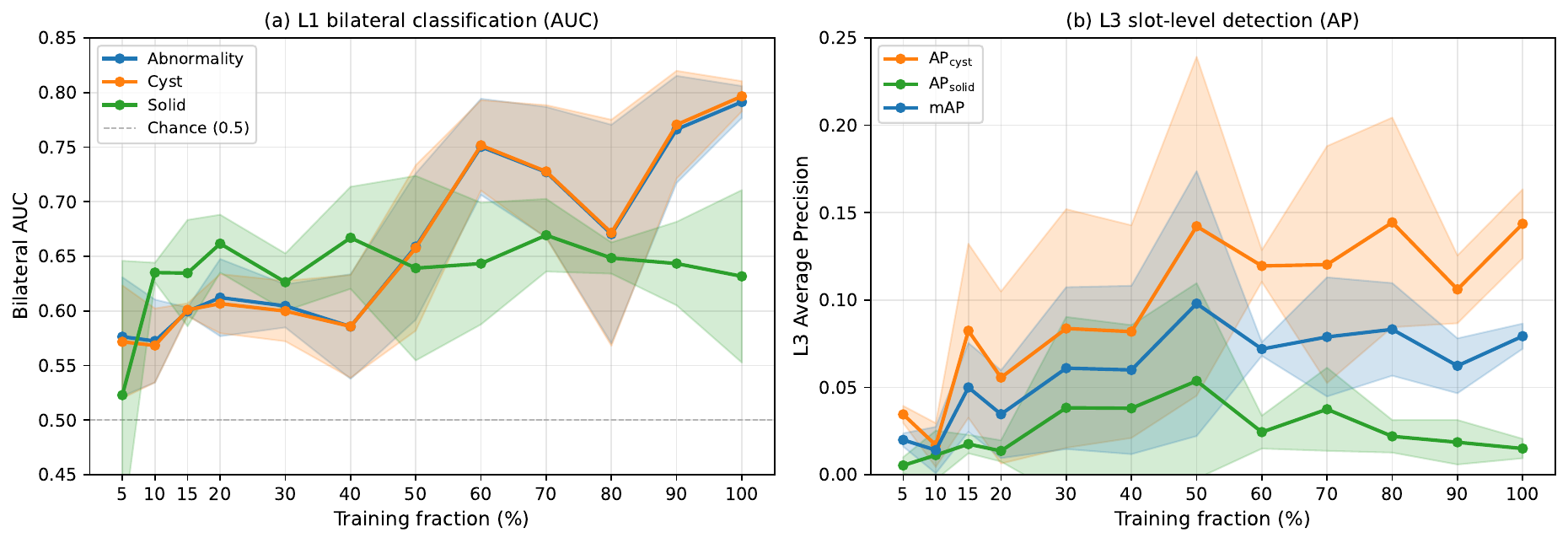}
    \caption{Data efficiency curves. \textbf{(a)} Side-level bilateral AUC for abnormality, cyst, and solid. \textbf{(b)} Per-lesion average precision (AP$_{\text{cyst}}$, AP$_{\text{solid}}$, mAP) at size tolerance $\tau = 1$~cm. Lines and shaded bands show mean and standard deviation across three seeds.}
    \label{fig:fewshot}
\end{figure}

Figure~\ref{fig:fewshot} shows how side-level and per-lesion metrics evolve with training-set size. We retrained the same configuration as our main experiments on subsets ranging from 5\% to 100\% of available exams, with three random seeds per fraction. At side-level (Panel~a), abnormality and cyst AUCs scale smoothly from around $0.55$ at the smallest fraction to approximately $0.80$ at full data, with incremental gains continuing throughout the full range. Solid classification at side-level is noisier but recoverable, ending around $0.65$ AUC at 100\%.

Per-lesion detection (Panel~b) shows the same qualitative scaling pattern at lower absolute levels. AP$_{\text{cyst}}$ rises from near zero at 5\% to approximately $0.15$ at 100\%, tracking the shape of cyst AUC at a substantially lower level. AP$_{\text{solid}}$ remains near the noise floor across all fractions, reflecting the scarcity of solid positives in our training set rather than a fundamental limit; lifting this curve likely requires more solid-class examples specifically rather than more data overall. Mean AP, dominated by AP$_{\text{cyst}}$, reaches approximately $0.08$ at 100\%. All three per-lesion curves are still trending upward at full training size, indicating that additional training data would continue to help common-class per-lesion detection, and that targeted solid collection would be needed for the rare class.
\FloatBarrier

\subsection{External validation on KiTS23}
\label{sec:results_kits}

To assess cross-dataset generalization, we evaluated our two main methods on KiTS23~\cite{myronenko2023kits}, an independent 489-case cohort with expert tumor and cyst segmentations. Side-level L1 and L2 labels were derived from these segmentations; although every patient has at least one confirmed tumor, roughly 30\% of individual kidney sides are unaffected, so side-level classification remains a meaningful task. KiTS23 also contains a much larger fraction of solid lesions than the UF cohort ($\sim$53\% vs $<$4\%). All volumes were processed through the same preprocessing pipeline, and no KiTS23 case was seen during training. A random-initialization baseline (untrained SwinUNETR + LesionDETR, three seeds) calibrates the zero-shot noise floor.

\begin{table}[!htbp]
\centering
\scriptsize
\setlength{\tabcolsep}{4pt}
\renewcommand{\arraystretch}{1.15}
\caption{Zero-shot evaluation on KiTS23 (inference only, no KiTS23 training exposure). Mean $\pm$ std over three seeds.}
\label{tab:kits}
\begin{tabular}{l l c c c c}
\toprule
\textbf{Method} & \textbf{Input} & \textbf{L1 abn.}\,$\uparrow$ & \textbf{L2 cyst}\,$\uparrow$ & \textbf{L2 solid}\,$\uparrow$ & \textbf{Size MAE (cm)}\,$\downarrow$ \\
\midrule
\multirow{2}{*}{LesionDETR + $L_1L_2$}
 & crop  & $0.770${\tiny$\pm 0.028$} & $\mathbf{0.776}${\tiny$\pm 0.021$} & $0.659${\tiny$\pm 0.025$} & $2.19${\tiny$\pm 0.07$} \\
 & fmask & $\mathbf{0.817}${\tiny$\pm 0.072$} & $0.755${\tiny$\pm 0.080$} & $\mathbf{0.670}${\tiny$\pm 0.049$} & $1.89${\tiny$\pm 0.14$} \\
\midrule
\multirow{2}{*}{Count-cond + focal}
 & crop  & $0.568${\tiny$\pm 0.044$} & $0.633${\tiny$\pm 0.059$} & $0.516${\tiny$\pm 0.016$} & $2.22${\tiny$\pm 0.19$} \\
 & fmask & $0.729${\tiny$\pm 0.070$} & $0.750${\tiny$\pm 0.026$} & $0.597${\tiny$\pm 0.066$} & $\mathbf{1.86}${\tiny$\pm 0.44$} \\
\midrule
Random initialization & ---   & $0.499${\tiny$\pm 0.029$} & $0.518${\tiny$\pm 0.034$} & $0.497${\tiny$\pm 0.012$} & $4.11${\tiny$\pm 0.08$} \\
\bottomrule
\end{tabular}
\end{table}

Table~\ref{tab:kits} shows that our UF-trained models transfer to KiTS23 despite no target-domain exposure. All four trained configurations exceed the random-initialization floor on every metric: side-level AUCs reach up to $0.817$ (L1), $0.776$ (cyst), and $0.670$ (solid) against floors near $0.5$, and per-side size MAE falls from the $4.11$~cm random floor to $1.86$--$2.22$~cm across all trained configurations, a roughly $50\%$ reduction.

\section{Discussion}
\label{sec:discussion}

\subsection{DETR-style detection and its trade-off}

Radiologists describe kidney CT findings one lesion at a time, whereas most prior computational work predicts at the patient or organ level. We formulate the task to match radiologists' practice. Our results show that DETR-style set prediction can be adapted to this task: LesionDETR reaches a stable side-level abnormality AUC of $0.799 \pm 0.009$, while the count-conditioned variant reaches mAP $0.190 \pm 0.036$ at the per-lesion level. To our knowledge, this is the first demonstration that DETR-style set prediction extends to 3D kidney lesion detection from radiology-report supervision.

The ablation further shows that no single head wins across all metrics: LesionDETR occupies the side-level end of a trade-off, while a count-conditioned variant occupies the per-lesion end. The latter reaches mAP $0.190 \pm 0.036$, roughly $2.5\times$ the random-classifier floor, at the cost of weaker side-level AUCs. These numbers indicate per-lesion detection is feasible in this setting, though absolute performance remains below clinical-deployment standards. The right head depends on the downstream use case: LesionDETR suits screening and triage workflows, while the count-conditioned head suits per-lesion structured-report pipelines.

The flatten-and-sort head's strength on the rare class (solid) is counterintuitive but reconciles with a small-sample argument: with only 20 solid positives in validation, Hungarian matching has few gradient examples to learn a reliable match cost, whereas size-based positional ordering provides a fixed structural prior that does not require training to converge. Size-sorted slot assignment may therefore outperform learned matching specifically when the class is rare, a regime where our DETR-style heads are data-starved. This advantage reaches significance relative to the learned-matching heads (Appendix~\ref{app:stats}) but not the summary baseline, consistent with the small-sample regime limiting the resolving power of the comparison.

\subsection{Design principles}

Several design choices affect performance. The largest single gain comes from providing a segmentation mask as an input channel: the full multi-class mask carries more noise than a binary kidney mask but yields additional gains. Hierarchical supervision is the second lever: aggregation connects per-slot predictions to side-level labels, letting a single L3-trained model also serve L1 and L2 evaluations, and adding L2 supervision on top of L1 improves side-level AUCs (LesionDETR $+0.087$ on L1 abnormality, $+0.105$ on L2 cyst) at modest cost to per-lesion mAP. The third lever is pretraining: SuPreM, which is pretrained on abdominal CT and thus shares the target domain, transfers most effectively among the six encoder initialisations we tested.
Two methodological choices support the reliability of these ablations. First, a layered learning rate, with the encoder trained more slowly than the detection head, helps preserve SuPreM's pretrained features during fine-tuning. Second, seed-to-seed variance can be large, making multi-seed evaluation necessary.

\subsection{Clinical workflow integration}
Consultations with our co-author urologists during framework design confirmed that clinicians value multi-granularity output over a single fixed level, since different clinical questions demand different levels of detail. The three-tier label hierarchy maps to these workflow stages: L1 can flag cases for radiologist attention, L2 supports clinical triage through cyst-versus-solid discrimination, and L3 supports structured reporting with per-lesion enhancement and attenuation attributes.
A second preference clinicians expressed is that per-lesion structured output is most useful as input to downstream report generation rather than as a standalone diagnostic tool.

\subsection{External validation on KiTS23}
\label{sec:discussion_kits}

The external evaluation confirms that the framework produces generalizable representations: all four trained configurations transfer to KiTS23 above the random-initialization floor despite no target-domain exposure.
The crop-vs-fmask gap on KiTS23 is much larger than it is in-domain ($0.05$--$0.16$ AUC vs $0.01$--$0.02$ AUC at side-level). The KiTS23 fmask uses expert voxel-level segmentations, whereas the UF fmask uses predictions from a separately trained nnUNet that carry some segmentation noise. The KiTS23 comparison therefore isolates a mask-quality lever that in-domain evaluation underestimates: accurate segmentation channels provide a real boost to downstream lesion characterization (up to $0.16$ AUC on abnormality). Upstream segmentation quality is thus a genuine bottleneck in the current framework, and improving it could close a meaningful fraction of the remaining room for improvement.

\subsection{Limitations: Data scale and clinical-threshold performance}

Data scale is the primary limitation of this study. Deep 3D models on medical CT are data-hungry, especially for rare classes such as solid lesions; our cohort of 2,619 volumes with 25 solid-class positives in validation is below the scale needed to push per-lesion detection to clinical-deployment performance. The training data is also drawn from a single academic medical center (UF Health), and external validation on KiTS23 is limited to inference without fine-tuning. Together, these factors explain the current performance ceiling: AP$_{\text{solid}}$ is at the random-baseline floor for both main configurations, and side-level abnormality AUC near $0.80$ leaves significant room for improvement before automated triage decisions could be safely made. Multi-center training data, ideally including tumor-enriched cohorts such as KiTS23 or surgical cohorts, would address both scale and generalizability. With only 20 solid positives in validation (13 left, 7 right), AP$_{\text{solid}}$ estimates carry sampling-induced variability on the order of $\pm 0.05$--$0.10$ (Appendix~\ref{app:random_baselines}); solid-class comparisons in this paper should be read as directional rather than definitive.

\subsection{Limitations: Reader study}

Clinicians were involved throughout the pipeline, including definition of the attribute schema. After discussions with our co-author urologists, we judged that a formal reader study is not currently needed: with the present per-lesion AP, such a study would conclude the model is not yet a usable clinical tool, which adds little beyond what our quantitative metrics already show. We plan to conduct a reader study once per-lesion performance crosses a clinically meaningful threshold.

\section{Conclusion}
\label{sec:conclusion}

We presented a systematic study of multi-granularity 3D kidney CT characterization. Our framework consists of three elements: a lesion-centric task formulation matching radiologists' practice, a dataset of 2,619 CT volumes paired with radiology reports, and LesionDETR, a DETR-style set-prediction architecture with size-distance Hungarian matching.

Ablations reveal a trade-off between side-level aggregation and per-lesion detection, with LesionDETR and a count-conditioned variant at opposite ends. Design levers include segmentation-mask inputs, hierarchical supervision, and same-domain pretraining. While per-lesion performance has not yet reached clinical-deployment thresholds, the framework establishes feasibility and provides a foundation for downstream structured reporting and report generation.

\section*{Acknowledgments}

Computational resources were provided by the University of Florida HiPerGator supercomputer.

\section*{Declaration of competing interest}

The authors declare that they have no known competing financial interests or personal relationships that could have appeared to influence the work reported in this paper.

\section*{Ethics statement}

This study was approved by the University of Florida Institutional Review Board (IRB202400720). All patient data were de-identified prior to analysis. Informed consent was waived due to the retrospective nature of the study.

\section*{Data availability}

The UF Health dataset is not publicly available due to patient privacy restrictions. The KiTS23 dataset is publicly available at \url{https://kits-challenge.org/kits23/}. Code will be made available upon acceptance.

\section*{Declaration of generative AI use}

Generative AI tools (Claude, Anthropic) were used to assist with literature search, code development, and manuscript drafting. All AI-generated content was reviewed, verified, and revised by the authors. The authors take full responsibility for the content of this publication.

\bibliographystyle{elsarticle-num}
\bibliography{references}

@article{silverman2019bosniak,
  author  = {Silverman, Stuart G. and Pedrosa, Ivan and Ellis, James H. and Hindman, Nicole M. and Schieda, Nicola and Smith, Andrew D. and Remer, Erick M. and Shinagare, Atul B. and Curci, Nicholas E. and Rader, Daniel J. and others},
  title   = {Bosniak Classification of Cystic Renal Masses, Version 2019: An Update Proposal and Needs Assessment},
  journal = {Radiology},
  volume  = {292},
  number  = {2},
  pages   = {475--488},
  year    = {2019},
}

@article{herts2018acr,
  author  = {Herts, Brian R. and Silverman, Stuart G. and Hindman, Nicole M. and Uzzo, Robert G. and Hartman, Robert P. and Israel, Gary M. and Baumgarten, Daniel A. and Sussman, Clara B. and Ramaiya, Nikhil H.},
  title   = {Management of the Incidental Renal Mass on {CT}: A White Paper of the {ACR} Incidental Findings Committee},
  journal = {Journal of the American College of Radiology},
  volume  = {15},
  number  = {2},
  pages   = {264--273},
  year    = {2018},
}

@article{lacpanet2024,
  author  = {Uhm, Kwang-Hyun and Jung, Seung-Won and Hong, Sung-Hoo and Ko, Sung-Jea},
  title   = {Lesion-Aware Cross-Phase Attention Network for Renal Tumor Subtype Classification on Multi-Phase {CT} Scans},
  journal = {Computers in Biology and Medicine},
  year    = {2024},
  note    = {arXiv:2406.16322},
}

@article{zhou2021endtoend,
  author  = {Zhou, Jianfei and others},
  title   = {An End-to-End Framework for Kidney Cancer Diagnosis from Multi-Phase {CT} Scans},
  journal = {npj Precision Oncology},
  volume  = {5},
  pages   = {24},
  year    = {2021},
}

@article{sung2021global,
  author  = {Sung, Hyuna and Ferlay, Jacques and Siegel, Rebecca L. and Laversanne, Mathieu and Soerjomataram, Isabelle and Jemal, Ahmedin and Bray, Freddie},
  title   = {Global Cancer Statistics 2020: {GLOBOCAN} Estimates of Incidence and Mortality Worldwide for 36 Cancers in 185 Countries},
  journal = {CA: A Cancer Journal for Clinicians},
  volume  = {71},
  number  = {3},
  pages   = {209--249},
  year    = {2021},
}

@article{isensee2021nnunet,
  author  = {Isensee, Fabian and Jaeger, Paul F. and Kohl, Simon A. A. and Petersen, Jens and Maier-Hein, Klaus H.},
  title   = {{nnU-Net}: A Self-Configuring Method for Deep Learning-Based Biomedical Image Segmentation},
  journal = {Nature Methods},
  volume  = {18},
  number  = {2},
  pages   = {203--211},
  year    = {2021},
}

@inproceedings{myronenko2023kits,
  author    = {Myronenko, Andriy and others},
  title     = {Automated {3D} Segmentation of Kidneys and Tumors in {MICCAI KiTS} 2023 Challenge},
  booktitle = {Springer LNCS},
  year      = {2023},
  note      = {arXiv:2310.04110},
}

@article{wasserthal2023totalseg,
  author  = {Wasserthal, Jakob and Breit, Hanns-Christian and Meyer, Manfred T. and others},
  title   = {{TotalSegmentator}: Robust Segmentation of 104 Anatomic Structures in {CT} Images},
  journal = {Radiology: Artificial Intelligence},
  volume  = {5},
  number  = {5},
  pages   = {e230024},
  year    = {2023},
}

@inproceedings{li2024suprem,
  author    = {Li, Wenxuan and others},
  title     = {How Well Do Supervised {3D} Models Transfer to Medical Imaging Tasks?},
  booktitle = {International Conference on Learning Representations (ICLR)},
  year      = {2024},
  note      = {Oral. GitHub: MrGiovanni/SuPreM},
}

@inproceedings{wu2024voco,
  author    = {Wu, Linshan and Zhuang, Jiaxin and others},
  title     = {{VoCo}: A Simple-yet-Effective Volume Contrastive Learning Framework for {3D} Medical Image Analysis},
  booktitle = {IEEE/CVF Conference on Computer Vision and Pattern Recognition (CVPR)},
  year      = {2024},
  note      = {arXiv:2402.17300. Extended version: IEEE TPAMI 2025. GitHub: Luffy503/VoCo},
}

@article{hamamci2024ctclip,
  author  = {Hamamci, Ibrahim Ethem and others},
  title   = {Developing Generalist Foundation Models from a Multimodal Dataset for {3D} Computed Tomography},
  journal = {Nature Biomedical Engineering},
  year    = {2025},
  note    = {arXiv:2403.17834},
}

@inproceedings{ctgraph2025,
  author    = {Kalisch, Hannah and others},
  title     = {{CT-GRAPH}: Hierarchical Graph Attention Network for Anatomy-Guided {CT} Report Generation},
  booktitle = {ICCV Workshop},
  year      = {2025},
  note      = {arXiv:2508.05375},
}

@inproceedings{baharoon2025design,
  author    = {Baharoon, Mohammed and others},
  title     = {Exploring the Design Space of {3D MLLMs} for {CT} Report Generation},
  booktitle = {Medical Image Computing and Computer-Assisted Intervention (MICCAI)},
  year      = {2025},
  note      = {arXiv:2506.21535},
}

@article{liang2025kidney,
  author = {Liang, Renjie and Fan, Zhengkang and Pan, Jinqian and Sun, Chenkun and Steinberg, Bruce Daniel and Terry, Russell and Xu, Jie},
  title  = {A Clinically-Grounded Two-Stage Framework for Renal {CT} Report Generation},
  year   = {2025},
  note   = {arXiv:2506.23584},
}

@inproceedings{bassi2025radgpt,
  author    = {Bassi, Pedro and others},
  title     = {{RadGPT}: Constructing {3D} Image-Text Tumor Datasets},
  booktitle = {IEEE/CVF International Conference on Computer Vision (ICCV)},
  year      = {2025},
  note      = {arXiv:2501.04678},
}

@inproceedings{tang2022swinunetr,
  author    = {Tang, Yucheng and others},
  title     = {Self-Supervised Pre-Training of {Swin Transformers} for {3D} Medical Image Analysis},
  booktitle = {IEEE/CVF Conference on Computer Vision and Pattern Recognition (CVPR)},
  year      = {2022},
  note      = {arXiv:2111.14791},
}

@article{huang2023stunet,
  author = {Huang, Ziyan and others},
  title  = {{STU-Net}: Scalable and Transferable Medical Image Segmentation Models Empowered by Large-Scale Supervised Pre-Training},
  year   = {2023},
  note   = {arXiv:2304.06716. GitHub: uni-medical/STU-Net},
}

@article{blankemeier2026merlin,
  author  = {Blankemeier, Louis and others},
  title   = {Merlin: A Computed Tomography Vision-Language Foundation Model and Dataset},
  journal = {Nature},
  year    = {2026},
  note    = {arXiv:2406.06512},
}

@article{bai2024m3d,
  author = {Bai, Fan and others},
  title  = {{M3D}: Advancing {3D} Medical Image Analysis with Multi-Modal Large Language Models},
  year   = {2024},
  note   = {arXiv:2404.00578},
}

@inproceedings{yan2019mulan,
  author    = {Yan, Ke and Tang, Youbao and Peng, Yifan and Sandfort, Veit and Bagheri, Mohammadhadi and Lu, Zhiyong and Summers, Ronald M.},
  title     = {{MULAN}: Multitask Universal Lesion Analysis Network for Joint Lesion Detection, Tagging, and Segmentation},
  booktitle = {Medical Image Computing and Computer-Assisted Intervention (MICCAI)},
  series    = {LNCS},
  volume    = {11769},
  pages     = {194--202},
  year      = {2019},
}

@inproceedings{zlocha2019improving,
  author    = {Zlocha, Martin and Dou, Qi and Glocker, Ben},
  title     = {Improving {RetinaNet} for {CT} Lesion Detection with Dense Masks from Weak {RECIST} Labels},
  booktitle = {Medical Image Computing and Computer-Assisted Intervention (MICCAI)},
  series    = {LNCS},
  volume    = {11769},
  pages     = {402--410},
  year      = {2019},
}

@inproceedings{oktay2018attention,
  author    = {Oktay, Ozan and Schlemper, Jo and Le Folgoc, Loic and Lee, Matthew and Heinrich, Mattias and Misawa, Kazunari and Mori, Kensaku and McDonagh, Steven and Hammerla, Nils Y. and Kainz, Bernhard and Glocker, Ben and Rueckert, Daniel},
  title     = {Attention {U-Net}: Learning Where to Look for the Pancreas},
  booktitle = {Medical Imaging with Deep Learning (MIDL)},
  year      = {2018},
  note      = {arXiv:1804.03999},
}

@article{kamnitsas2017efficient,
  author  = {Kamnitsas, Konstantinos and Ledig, Christian and Newcombe, Virginia F. J. and Simpson, Joanna P. and Kane, Andrew D. and Menon, David K. and Rueckert, Daniel and Glocker, Ben},
  title   = {Efficient Multi-Scale {3D CNN} with Fully Connected {CRF} for Accurate Brain Lesion Segmentation},
  journal = {Medical Image Analysis},
  volume  = {36},
  pages   = {61--78},
  year    = {2017},
}

@article{xi2020stratification,
  author  = {Xi, Ianto Lin and Zhao, Yijun and Fishman, Ronald and Kamyab, Michael and Lipton, Paul and Lehrer, Brendon and Hao, Minghao and Raman, Steven S.},
  title   = {Stratification of Cystic Renal Masses into Benign and Potentially Malignant: Applying Machine Learning to the {Bosniak} Classification},
  journal = {European Radiology},
  volume  = {30},
  pages   = {2817--2826},
  year    = {2020},
}

@article{han2019classification,
  author  = {Han, Seokmin and Hwang, Sung Il and Lee, Hak Jong},
  title   = {The Classification of Renal Cancer in 3-Phase {CT} Images Using a Deep Learning Method},
  journal = {Journal of Digital Imaging},
  volume  = {32},
  number  = {4},
  pages   = {638--643},
  year    = {2019},
}

@article{uhlig2022deep,
  author  = {Uhlig, Johannes and Parakh, Anushri and Sauer, Bettina and others},
  title   = {Deep Learning and Radiomic Feature-Based Blending Ensemble Classifier for Malignancy Risk Prediction in Cystic Renal Lesions},
  journal = {Insights into Imaging},
  volume  = {13},
  pages   = {6},
  year    = {2022},
}

@article{klontzas2024cnn,
  author  = {Klontzas, Michail E. and Kalarakis, Georgios and Koltsakis, Emmanouil and Papathomas, Thomas and Karantanas, Apostolos H. and Tzortzakakis, Antonios},
  title   = {Convolutional Neural Networks for the Differentiation between Benign and Malignant Renal Tumors with a Multicenter International Computed Tomography Dataset},
  journal = {Insights into Imaging},
  volume  = {15},
  pages   = {48},
  year    = {2024},
}

@inproceedings{lin2017focal,
  author    = {Lin, Tsung-Yi and Goyal, Priya and Girshick, Ross and He, Kaiming and Doll{\'a}r, Piotr},
  title     = {Focal Loss for Dense Object Detection},
  booktitle = {IEEE International Conference on Computer Vision (ICCV)},
  pages     = {2980--2988},
  year      = {2017},
}

@article{adams2023leveraging,
  author  = {Adams, Lisa C. and Truhn, Daniel and Busch, Felix and Kader, Avan and Niehues, Stefan M. and Makowski, Marcus R. and Bressem, Keno K.},
  title   = {Leveraging {GPT-4} for Post Hoc Transformation of Free-Text Radiology Reports into Structured Reporting: A Multilingual Feasibility Study},
  journal = {Radiology},
  volume  = {307},
  number  = {4},
  pages   = {e230725},
  year    = {2023},
}

@inproceedings{irvin2019chexpert,
  author    = {Irvin, Jeremy and Rajpurkar, Pranav and Ko, Michael and Yu, Yifan and Ciurea-Ilcus, Silviana and Chute, Chris and Marklund, Henrik and Haghgoo, Behzad and Ball, Robyn and Shpanskaya, Katie and Seekins, Jayne and Mong, David A. and Halabi, Safwan S. and Sandberg, Jesse K. and Jones, Ricky and Larson, David B. and Langlotz, Curtis P. and Patel, Bhavik N. and Lungren, Matthew P. and Ng, Andrew Y.},
  title     = {{CheXpert}: A Large Chest Radiograph Dataset with Uncertainty Labels and Expert Comparison},
  booktitle = {AAAI Conference on Artificial Intelligence},
  volume    = {33},
  number    = {01},
  pages     = {590--597},
  year      = {2019},
}

@article{tajbakhsh2016cnn,
  author  = {Tajbakhsh, Nima and Shin, Jae Y. and Gurudu, Suryakanth R. and Hurst, R. Todd and Kendall, Christopher B. and Gotway, Michael B. and Liang, Jianming},
  title   = {Convolutional Neural Networks for Medical Image Analysis: Full Training or Fine Tuning?},
  journal = {IEEE Transactions on Medical Imaging},
  volume  = {35},
  number  = {5},
  pages   = {1299--1312},
  year    = {2016},
}

@inproceedings{raghu2019transfusion,
  author    = {Raghu, Maithra and Zhang, Chiyuan and Kleinberg, Jon and Bengio, Samy},
  title     = {Transfusion: Understanding Transfer Learning for Medical Imaging},
  booktitle = {Advances in Neural Information Processing Systems (NeurIPS)},
  year      = {2019},
}

@article{zhou2021models,
  author  = {Zhou, Zongwei and Sodha, Vatsal and Pang, Jiaxuan and Gotway, Michael B. and Liang, Jianming},
  title   = {Models Genesis},
  journal = {Medical Image Analysis},
  volume  = {67},
  pages   = {101840},
  year    = {2021},
}

@article{zhao2023multitask,
  author  = {Zhao, Yan and Wang, Xiuying and Che, Tongtong and Bao, Guoqing and Li, Shuyu},
  title   = {Multi-Task Deep Learning for Medical Image Computing and Analysis: A Review},
  journal = {Computers in Biology and Medicine},
  volume  = {153},
  pages   = {106496},
  year    = {2023},
}

@inproceedings{carion2020detr,
  author    = {Carion, Nicolas and Massa, Francisco and Synnaeve, Gabriel and Usunier, Nicolas and Kirillov, Alexander and Zagoruyko, Sergey},
  title     = {End-to-End Object Detection with Transformers},
  booktitle = {European Conference on Computer Vision (ECCV)},
  series    = {LNCS},
  volume    = {12346},
  pages     = {213--229},
  year      = {2020},
}

@article{li2025detr_medical,
  author  = {Li, Huazhang and others},
  title   = {Transformer-Powered Precision: A {DETR}-Based Approach for Robust Detection in Medical Ultrasound with Cholelithiasis as a Case Study},
  journal = {Computational and Structural Biotechnology Journal},
  year    = {2025},
}

@inproceedings{lin2014coco,
  author    = {Lin, Tsung-Yi and Maire, Michael and Belongie, Serge and Hays, James and Perona, Pietro and Ramanan, Deva and Doll{\'a}r, Piotr and Zitnick, C. Lawrence},
  title     = {Microsoft {COCO}: Common Objects in Context},
  booktitle = {European Conference on Computer Vision (ECCV)},
  pages     = {740--755},
  year      = {2014},
}

\newpage
\appendix

\section{Representative report snippets}
\label{app:terminology}

\noindent\fcolorbox{gray}{gray!15}{\parbox{0.97\linewidth}{%
\footnotesize
\textbf{Sample 1.}\\
\textit{FINDINGS:} Kidneys: Right kidney: No hydronephrosis. No obstructing calculus. There is a mildly lobular, solid heterogeneously enhancing right renal mass with additional characteristics as follows: Location: Right upper thirds posterolaterally, endophytic. Mass insinuates into the upper and mid calyces. Size: Approximately 6.5 cm transverse by 5.0 cm AP by 5.7 cm craniocaudad. Enhancement pattern: Brisk heterogeneous enhancement with scattered areas of central necrosis. Capsule: Appears to be intact for the most part. Left kidney: Normal enhancement. No hydronephrosis. At least 2 small cysts, dominant cyst is in the superior pole, measuring approximately 0.9~cm in diameter. In addition, there is nonobstructing 3~mm calculus in an upper pole calyx. No solid masses seen.
}}
\vspace{0.5em}

\noindent\fcolorbox{gray}{gray!15}{\parbox{0.97\linewidth}{%
\footnotesize
\textbf{Sample 2.}\\
\textit{FINDINGS:} Kidneys, Ureters \& Collecting System: 3.4~cm right renal cyst measures fluid attenuation with similar 3.3~cm cyst of the left inferior pole, with multiple additional smaller cysts. There are scattered \textbf{subcentimeter} low attenuating lesions that are too small to characterize, but likely benign. 3~mm calculus of the inferior left renal calyx is noted. No hydronephrosis or enhancing lesion bilaterally.
}}
\vspace{0.5em}

\noindent\fcolorbox{gray}{gray!15}{\parbox{0.97\linewidth}{%
\footnotesize
\textbf{Sample 3.}\\
\textit{FINDINGS:} Kidneys: Both kidneys enhance homogeneously. There is a 2.2~$\times$~1.5~cm complex cystic lesion with multiple thin septa in the superior pole. Few additional \textbf{subcentimeter} low attenuating lesions in both kidneys are too small to characterize but likely benign. There is a 3~mm nonobstructing calculus in the interpolar calyx of the left kidney.
}}
\vspace{0.5em}

To ground the label-schema decisions described in Section~\ref{sec:labels}, we present three representative snippets from radiology reports in our UF Health cohort. Patient identifiers and dates have been redacted; otherwise the text is verbatim.
These snippets illustrate four aspects of the source data that shaped our label schema:
\begin{enumerate}
\item \textit{Multi-lesion reports are common.} Samples~1 and~2 each describe three or more lesions of mixed type across both kidneys. A side-level summary would lose this structure, so we predict attributes independently for each lesion rather than aggregating across a kidney.
\item \textit{Attribute coverage is non-uniform.} In Sample~1, the right-kidney mass is described with five attributes (location, size, enhancement, capsule status, calyceal involvement), while the left-kidney cysts have only a count and a dominant size. Sample~2 reports two sized cysts alongside ``multiple additional smaller cysts'' with no individual attributes.
\item \textit{Qualitative and numerical sizes coexist.} Samples~2 and~3 mix the qualitative descriptor ``subcentimeter'' (mapped to 0.8~cm) with specific sizes such as ``2.2 $\times$ 1.5~cm''.
\item \textit{Mass and tumor are interchangeable.} In our cohort, 45 of 47 tumor lesions also carry the mass flag (Table~\ref{tab:label_dist}). We therefore merge mass and tumor into a single \emph{solid} label.
\end{enumerate}

\section{Lesion-Level Evaluation Protocol Validation}
\label{app:eval_validation}

Because the evaluation protocol is new, we validated both implementation correctness and AP sensitivity to changes in prediction quality.

\subsection{Golden test cases}
\label{app:eval_golden}

Table~\ref{tab:app_eval_golden} lists seven golden test cases. Edge cases (A, D, F) verify trivial corners: perfect predictions yield AP=1, an empty class is undefined, and all detections beyond the size tolerance yield AP=0. The four partial-failure modes are more diagnostic. In B, all FPs precede all TPs in ranking, yielding AP $=$ TP/(TP+FP) $= 0.333$. In C, half recall with no FPs achieves precision 1 but caps AP at $6/11 \approx 0.545$ under 11-point Pascal interpolation. E (random predictions on sparse GT) reaches approximately chance-level AP ($\approx 0.10$). G (ambiguous type scores at 0.5) still achieves AP $= 1.0$ because each patient has only one prediction, so no FP competes for ranking; this illustrates AP's invariance to confidence magnitude when ranking is correct. 

\begin{table}[!htbp]
\centering
\scriptsize
\setlength{\tabcolsep}{5pt}
\renewcommand{\arraystretch}{1.15}
\caption{Golden test cases for the per-class AP implementation. \emph{All cases pass}.}
\label{tab:app_eval_golden}
\begin{tabular}{l l c c}
\toprule
\textbf{Scenario} & \textbf{Construction} & \textbf{Expected AP} & \textbf{Computed} \\
\midrule
\multicolumn{4}{l}{\textit{Edge cases}} \\
A.\ Perfect          & All GTs detected, no FPs           & $1.000$ & $1.000$ \\
D.\ Empty class      & $n_{pos}=0$                         & $\text{NaN}$ & $\text{NaN}$ \\
F.\ Size beyond tol.\ & All matches fail size threshold   & $0.000$ & $0.000$ \\
\midrule
\multicolumn{4}{l}{\textit{Partial-failure modes}} \\
B.\ Inverted conf    & TPs ranked below FPs                & $0.333$ & $0.333$ \\
C.\ Half recall      & 50\% recall, no FPs                 & $0.545$ & $0.545$ \\
E.\ Random           & Random predictions, sparse GT       & ${\approx}0.10$ & $0.094$ \\
G.\ Ambiguous type   & All type scores at $0.5$            & $1.000$ & $1.000$ \\
\bottomrule
\end{tabular}
\end{table}

\subsection{Sensitivity analysis}
\label{app:eval_sensitivity}

\begin{figure}[!htbp]
\centering
\includegraphics[width=0.92\textwidth]{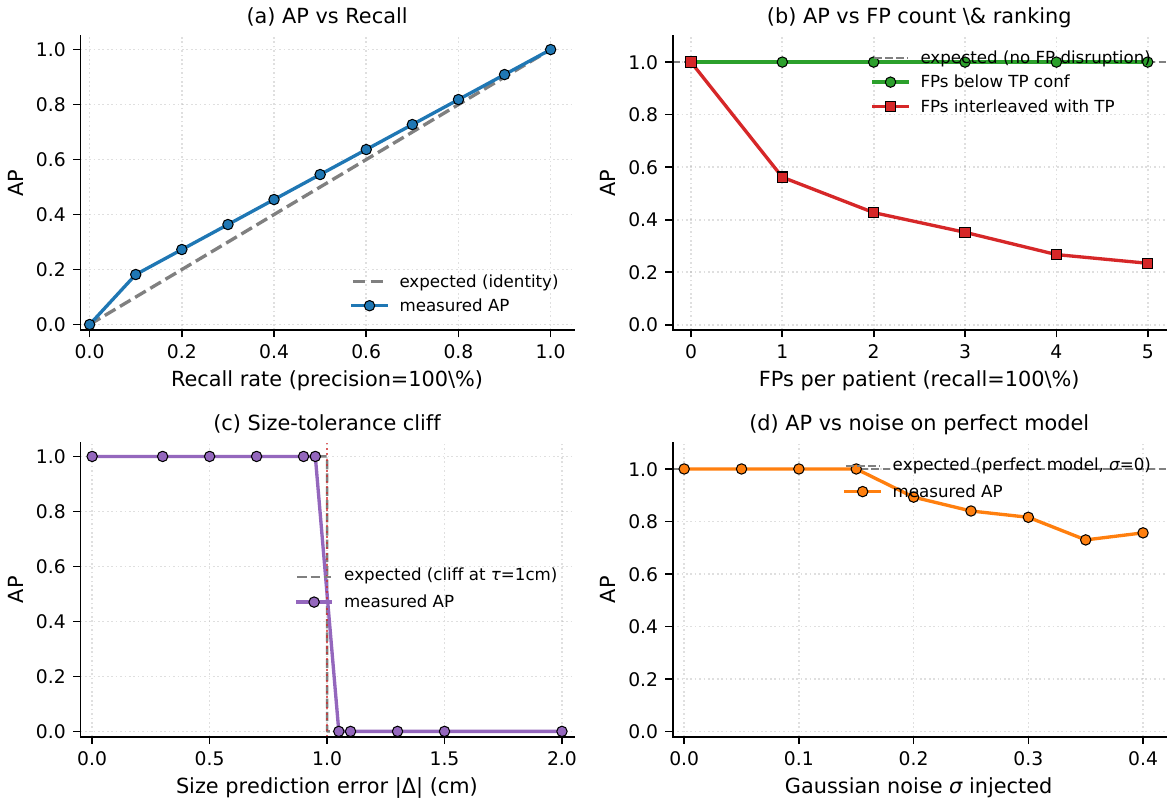}
\caption{Per-class AP sensitivity across four dimensions. \textbf{(a)} Recall sweep at precision=100\%. \textbf{(b)} FP count with TP and FP confidences uniformly interleaved in rank (recall=100\%). \textbf{(c)} Size-prediction error at the $\tau=1$\,cm tolerance. \textbf{(d)} Gaussian noise $\sigma$ injected into a perfect model. Dashed: theoretical expected AP; solid: measured AP.}
\label{fig:app_eval_sensitivity}
\end{figure}

Figure~\ref{fig:app_eval_sensitivity} sweeps four performance dimensions.
\textbf{(a) Recall.} Precision fixed at 100\%, recall swept. AP closely tracks the identity line, so recall is the primary AP ceiling.
\textbf{(b) FP ranking.} FP confidences interleaved among TP confidences (adversarial ranking). Measured AP tracks the theoretical decay $1/(1+\mathrm{FP})$; FPs ranked below all TPs leave AP unchanged (not shown). AP depends on FP ranking, not raw count.
\textbf{(c) Size error.} A uniform size error applied to every prediction. The $\tau = 1$~cm threshold is a binary cliff with no soft transition.
\textbf{(d) Noise.} Gaussian noise $\sigma$ injected on a perfect model. AP remains above $0.95$ for $\sigma < 0.2$ and degrades gradually beyond, supporting its use as a smooth training metric.
All four responses match theoretical predictions of 11-point interpolated AP. AP estimates are nevertheless unstable at low class prevalence: at $n_{pos} \le 5$, standard errors exceed $0.10$ and single rank flips can shift AP by more than $0.2$, which is why we treat rare-class AP as directional rather than definitive.

\section{Complete per-attribute results for the core matrix}
\label{app:core_matrix_full}

Table~\ref{tab:supp_core} reports per-attribute metrics not included in Table~\ref{tab:corematrix}: L2 solid AUC with left/right decomposition, and L3-only enhancement/attenuation accuracy and per-side count error. L3 rows use LesionDETR with hierarchical $L_1L_2$ supervision. Solid columns have small validation sets (L: $n = 13$, R: $n = 7$) and should be read as directional.

\begin{table}[!htbp]
\centering
\scriptsize
\setlength{\tabcolsep}{5pt}
\renewcommand{\arraystretch}{1.15}
\caption{Supplementary per-attribute metrics complementing Table~\ref{tab:corematrix}.}
\label{tab:supp_core}
\begin{tabular}{l cc c c c}
\toprule
\textbf{CT Input} & \textbf{Left solid}\,$\uparrow$ & \textbf{Right solid}\,$\uparrow$ & \textbf{Enh acc}\,$\uparrow$ & \textbf{Att acc}\,$\uparrow$ & \textbf{Count MAE}\,$\downarrow$ \\
\midrule
\multicolumn{6}{l}{\textit{Trained at L2 (lesion typing and size)}} \\
Whole CT      & 0.683 & \textbf{0.909} & --- & --- & --- \\
Cropped CT    & 0.399 & 0.891 & --- & --- & --- \\
+ Kidney mask & 0.430 & 0.893 & --- & --- & --- \\
+ Full mask   & 0.530 & 0.877 & --- & --- & --- \\
\midrule
\multicolumn{6}{l}{\textit{Trained at L3 (per-lesion detection)}} \\
Whole CT      & 0.543 & 0.826 & 0.525 & \textbf{0.856} & \textbf{1.139} \\
Cropped CT    & \textbf{0.565} & 0.833 & 0.496 & 0.498 & 1.218 \\
+ Kidney mask & 0.476 & \textbf{0.845} & \textbf{0.564} & 0.750 & 1.278 \\
+ Full mask   & 0.542 & 0.821 & 0.475 & 0.705 & 1.664 \\
\bottomrule
\end{tabular}
\end{table}

\section{Random baseline noise floors for per-lesion AP}
\label{app:random_baselines}

Per-lesion Average Precision on a small-cohort rare class can take sizeable non-zero values even under random inputs. To establish noise floors against which absolute AP numbers should be read, we compute four reference baselines using the same evaluation protocol; only the scoring function differs: \emph{full random} uses uniformly random slot fields; \emph{half random} keeps size from a trained model but randomizes class confidence; \emph{size only} sets confidence to the normalized predicted lesion size; and \emph{random init} is a single forward pass through an untrained SwinUNETR encoder feeding a randomly initialized LesionDETR. \emph{random init} sets the most conservative floor. Our per-lesion method (Count-cond) clears the AP$_{\text{cyst}}$ floor by roughly $2\times$, confirming that its AP reflects learned signal rather than architectural prior.

\begin{table}[!htbp]
\centering
\footnotesize
\setlength{\tabcolsep}{5pt}
\renewcommand{\arraystretch}{1.15}
\caption{Four random baseline noise floors.}
\label{tab:random_baselines}
\begin{tabular}{l c c c}
\toprule
\textbf{Baseline / Method} & AP$_{\text{cyst}}$ & AP$_{\text{solid}}$ & mAP \\
\midrule
full random & $0.116$ & $0.010$ & $0.063$ \\
half random & $0.141$ & $0.014$ & $0.077$ \\
size only   & $0.171$ & $0.005$ & $0.088$ \\
random init & $0.203$ & $0.001$ & $0.102$ \\
\midrule
Count-cond (ours) & $0.369$ & $0.012$ & $0.190$ \\
\bottomrule
\end{tabular}
\end{table}

\section{Loss-modifier and architecture ablation}
\label{app:loss_modifier_ablation}

\begin{figure}[!htbp]
\centering
\includegraphics[width=0.95\textwidth]{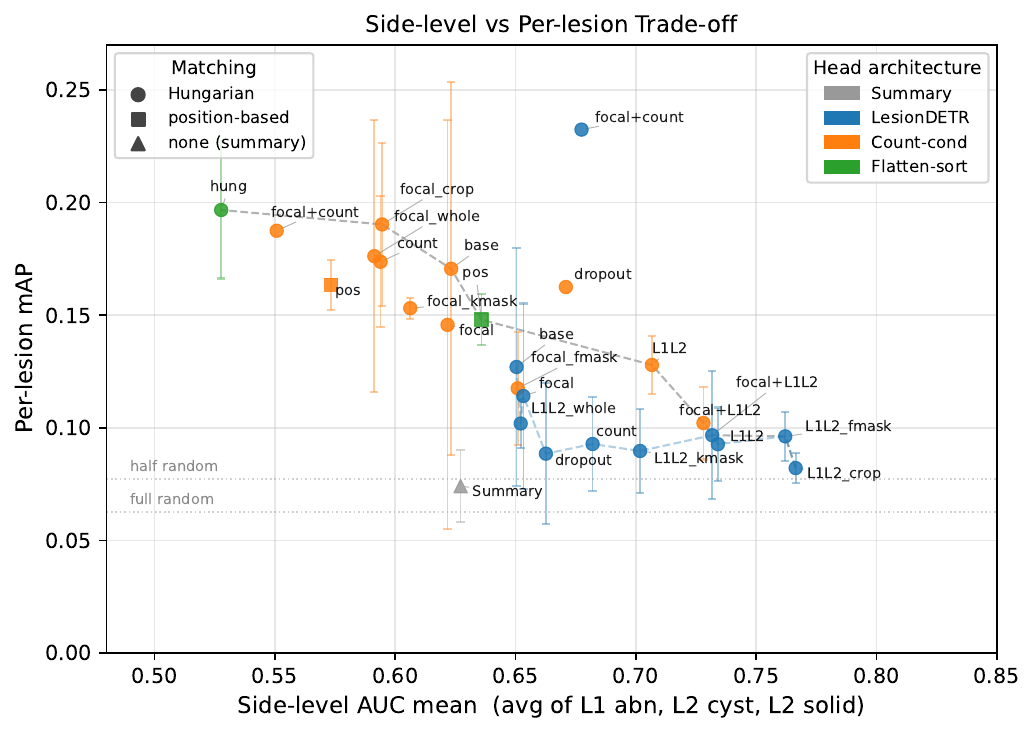}
\caption{Side-level vs.\ per-lesion trade-off across all head architectures and loss-modifier variants. Color denotes head architecture; marker shape denotes matching mechanism. Error bars are for three seeds. Horizontal dotted lines are random-baseline mAP floors.}
\label{fig:pareto}
\end{figure}

Figure~\ref{fig:pareto} visualizes the side-level-vs-per-lesion trade-off across four head architectures (LesionDETR, count-conditioned, flatten-and-sort, summary) and four standard loss modifiers applied to LesionDETR and Count-cond: focal loss~\cite{lin2017focal} on the existence logit, a count regulariser on the per-side predicted lesion count, hierarchical $L_1L_2$ aggregation supervising side-level abnormality and per-type presence, and dropout on the prediction head. Head architecture has a much larger effect on the metric-family ranking than any loss-modifier within a head: LesionDETR variants cluster tightly at the side-level-dominant end of the frontier (mean AUC $0.65$--$0.77$, mAP $0.082$--$0.127$), while count-conditioned variants cluster at the per-lesion end (mAP $0.102$--$0.190$, mean AUC $0.59$--$0.73$).

\section{Layer-wise linear probing of frozen encoders}
\label{app:probing}

To localize where SuPreM's advantage originates within the encoder, we probe the frozen representations at each depth of the SwinUNETR backbone for the three variants: SuPreM, VoCo, and a randomly initialized baseline. For each encoder stage we extract the intermediate feature map, apply global average pooling, and fit a logistic regression on the pooled vector to predict side-level L1 abnormality; the encoder weights are never updated.

\begin{table}[!htbp]
\centering
\scriptsize
\setlength{\tabcolsep}{8pt}
\renewcommand{\arraystretch}{1.1}
\caption{Layer-wise linear probing AUC. Stage 0 is the shallowest; stage 4 is the deepest. \textit{Fine-tuned}: end-to-end fine-tuning upper bound for SuPreM with full mask input at L1 (bilateral average from Table~\ref{tab:corematrix}).}
\label{tab:probing}
\begin{tabular}{lccccc c}
\toprule
\textbf{Encoder} & \multicolumn{5}{c}{\textbf{Linear probing}} & \textbf{Fine-tuned} \\
\cmidrule(lr){2-6}
 & Stage 0 & Stage 1 & Stage 2 & Stage 3 & Stage 4 & \\
\midrule
SwinUNETR scratch   & 0.508 & 0.502 & 0.507 & 0.544 & 0.512 & --- \\
VoCo                & 0.506 & 0.538 & 0.556 & 0.540 & 0.494 & --- \\
\textbf{SuPreM}     & 0.482 & 0.583 & 0.591 & 0.605 & 0.570 & \textbf{0.840} \\
\bottomrule
\end{tabular}
\end{table}

Two patterns stand out. First, SuPreM's benefit over the random baseline is concentrated at mid-to-deep layers, not at shallow ones: the two encoders are tied at the shallowest stage, and the gap opens up from stage~1 and peaks at stage~3. This concentration at mid-to-deep layers is consistent with SuPreM's fine-tuned advantage over scratch. Second, the best probing AUC (0.605 at SuPreM stage~3) remains far below the end-to-end fine-tuned upper bound (0.840), showing that a significant part of SuPreM's transferable signal is unlocked only through task-specific fine-tuning and cannot be recovered by a linear classifier on frozen features. We therefore fine-tune the encoder end-to-end rather than use SuPreM as a frozen feature extractor.

\section{Statistical comparisons}
\label{app:stats}

Pairwise comparisons between key configurations are assessed via paired patient-level bootstrap. For each pair $(A, B)$ and each of $1{,}000$ iterations, we resample validation patients with replacement and compute the metric on the same resampled patient set for both methods, averaged across the three training seeds. We report the point estimates for $A$ and $B$, the mean paired difference $\Delta = A - B$, the $95\%$ confidence interval (2.5th--97.5th percentiles of the bootstrap distribution of $\Delta$), and a one-sided $p$-value defined as $P[\Delta \le 0]$.

\begin{table}[!htbp]
\centering
\scriptsize
\setlength{\tabcolsep}{4pt}
\renewcommand{\arraystretch}{1.15}
\caption{Paired bootstrap significance tests for pairwise method comparisons. $\Delta = A - B$; $p$ is one-sided.}
\label{tab:stats}
\begin{threeparttable}
\begin{tabular}{l l c c r l c}
\toprule
\textbf{Method A} & \textbf{Method B} & \textbf{A} & \textbf{B} & \multicolumn{1}{c}{$\boldsymbol{\Delta}$} & \textbf{95\% CI} & \textbf{\textit{p}} \\
\midrule
\multicolumn{7}{l}{\textit{UF validation --- bilateral L1 abnormality AUC}} \\
LesionDETR & Count-cond & $0.803$ & $0.676$ & $+0.128$ & $[+0.100,\,+0.160]$ & $<0.001$ \\
LesionDETR & LesionDETR\tnote{a} & $0.803$ & $0.640$ & $+0.164$ & $[+0.131,\,+0.199]$ & $<0.001$ \\
Count-cond & Count-cond\tnote{a} & $0.676$ & $0.590$ & $+0.085$ & $[+0.043,\,+0.124]$ & $<0.001$ \\
SuPreM\tnote{b} & SwinUNETR scratch\tnote{b} & $0.781$ & $0.540$ & $+0.241$ & $[+0.211,\,+0.271]$ & $<0.001$ \\
\midrule
\multicolumn{7}{l}{\textit{UF validation --- per-lesion mAP}} \\
Count-cond & LesionDETR & $0.190$ & $0.082$ & $+0.107$ & $[+0.087,\,+0.127]$ & $<0.001$ \\
Count-cond & Random init & $0.190$ & $0.102$ & $+0.107$ & $[+0.076,\,+0.142]$ & $<0.001$ \\
\midrule
\multicolumn{7}{l}{\textit{UF validation --- per-class $\mathrm{AP}_{\text{solid}}$}} \\
Flatten-sort & LesionDETR & $0.109$ & $0.020$ & $+0.076$ & $[+0.005,\,+0.157]$ & $0.009$ \\
Flatten-sort & Count-cond & $0.109$ & $0.012$ & $+0.084$ & $[+0.011,\,+0.168]$ & $0.001$ \\
Flatten-sort & Summary head & $0.109$ & $0.076$ & $+0.040$ & $[-0.049,\,+0.133]$ & $0.217$ \\
\midrule
\multicolumn{7}{l}{\textit{KiTS23 external validation --- bilateral L1 abnormality AUC}} \\
LesionDETR\tnote{c} & Random init & $0.814$ & $0.499$ & $+0.316$ & $[+0.293,\,+0.339]$ & $<0.001$ \\
Count-cond\tnote{c} & Random init & $0.721$ & $0.499$ & $+0.223$ & $[+0.197,\,+0.249]$ & $<0.001$ \\
\bottomrule
\end{tabular}
\begin{tablenotes}[flushleft]
\footnotesize
\item[] $^a$ no hierarchical supervision; $^b$ kidney mask input; $^c$ full mask input.
\end{tablenotes}
\end{threeparttable}
\end{table}

All comparisons supporting main-text claims are statistically distinguishable ($p < 0.001$ for most, $p = 0.009$ for the Flatten-sort versus LesionDETR comparison on AP$_{\text{solid}}$). The exception is Flatten-sort versus the summary head on AP$_{\text{solid}}$: the $+0.040$ advantage is not significant ($p = 0.217$), so Flatten-sort's rare-class surprise holds against learned-matching heads but not the no-slot baseline. Encoder pretraining is the largest single lever: SuPreM vs SwinUNETR scratch, $\Delta = +0.241$ mean AUC.

\end{document}